\documentclass[acmsmall]{acmart}

\usepackage{geometry}
\usepackage{makecell}
\usepackage{multirow}
\usepackage{multicol}
\usepackage{array}
\usepackage{booktabs}
\usepackage{adjustbox}
\usepackage{algorithm}
\usepackage{algpseudocode}
\usepackage{tabularx}
\usepackage{tcolorbox}
\usepackage{colortbl}
\usepackage{graphicx}
\usepackage{longtable}
\usepackage{xcolor}
\usepackage{titlesec}

\AtBeginDocument{%
  }

\setcopyright{rightsretained}
\acmYear{2024}
\acmDOI{10.1145/3698145}
\acmJournal{PACMHCI}
\acmVolume{8}
\acmNumber{ISS}
\acmArticle{545}
\acmMonth{12}
\received{2024-07-01}
\received[accepted]{2024-09-20}

\begin{document}

\title{GestureGPT: Toward Zero-Shot Free-Form Hand Gesture Understanding with Large Language Model Agents}


\author{Xin Zeng}
\authornote{Both authors contributed equally to this research.}
\orcid{0009-0004-9862-4168}
\affiliation{%
  \institution{Institute of Computing Technology, Chinese Academy of Sciences}
  \city{Beijing}
  \country{China}
}
\affiliation{%
  \institution{University of Chinese Academy of Sciences}
  \city{Beijing}
  \country{China}
}
\email{zengxin18z@ict.ac.cn}

\author{Xiaoyu Wang}
\authornotemark[1]
\authornote{This work was completed during an internship at ICT, CAS.}
\orcid{0009-0007-6812-7031}
\affiliation{%
  \institution{The Hong Kong University of Science and Technology}
  \city{Hong Kong}
  \country{China}
}
\email{xwangij@connect.ust.hk}

\author{Tengxiang Zhang}
\authornote{Corresponding author. Part of the work was conducted while the author was at ICT, CAS.}
\orcid{0000-0002-0949-2801}
\affiliation{%
  \institution{Goertek Inc.}
  \city{Beijing}
  \country{China}
}
\email{ztxseuthu@gmail.com}

\author{Chun Yu}
\orcid{0000-0003-2591-7993}
\affiliation{%
  \institution{Computer Science and Technology, Tsinghua University}
  \city{Beijing}
  \country{China}
}

\author{Shengdong Zhao}
\orcid{0000-0001-7971-3107}
\affiliation{%
  \institution{Synteraction Lab, City University of Hong Kong}
  \city{Hong Kong}
  \country{China}
}

\author{Yiqiang Chen}
\orcid{0000-0002-8407-0780}
\authornote{Corresponding author.}
\affiliation{%
  \institution{Institute of Computing Technology, Chinese Academy of Sciences}
  \city{Beijing}
  \country{China}
}
\affiliation{%
  \institution{University of Chinese Academy of Sciences}
  \city{Beijing}
  \country{China}
}
\email{yqchen@ict.ac.cn}


\newcommand*{\eg}{\textit{e.g.},\;}
\newcommand*{\ie}{\textit{i.e.},\;}
\newcommand*{\vs}{\textit{v.s.}\;}
\newcommand*{\etc}{\textit{etc.}}
\newcommand*{\st}{\textit{s.t.},\;}
\newcommand*{\etal}{\textit{et~al.}\;}


\begin{abstract}
  Existing gesture interfaces only work with a fixed set of gestures defined either by interface designers or by users themselves, which introduces learning or demonstration efforts that diminish their naturalness.
  Humans, on the other hand, understand free-form gestures by synthesizing the gesture, context, experience, and common sense.
  In this way, the user does not need to learn, demonstrate, or associate gestures.
  We introduce GestureGPT, a free-form hand gesture understanding framework that mimics human gesture understanding procedures to enable a natural free-form gestural interface.
  Our framework leverages multiple Large Language Model agents to manage and synthesize gesture and context information, then infers the interaction intent by associating the gesture with an interface function. 
  More specifically, our triple-agent framework includes a Gesture Description Agent that automatically segments and formulates natural language descriptions of hand poses and movements based on hand landmark coordinates.
  The description is deciphered by a Gesture Inference Agent through self-reasoning and querying about the interaction context (\eg interaction history, gaze data), which is managed by a Context Management Agent.
  Following iterative exchanges, the Gesture Inference Agent discerns the user's intent by grounding it to an interactive function.
  We validated our framework offline under two real-world scenarios: smart home control and online video streaming.
  The average zero-shot Top-1/Top-5 grounding accuracies are 44.79\%/83.59\% for smart home tasks and 37.50\%/73.44\% for video streaming tasks.
  We also provide an extensive discussion that includes rationale for model selection, generalizability, and future research directions for a practical system \etc
\end{abstract}



\definecolor{description}{HTML}{70adde}
\definecolor{inference}{HTML}{e6c35d}
\definecolor{context}{HTML}{429984}

\begin{teaserfigure}
  \centering
  \includegraphics[width=\textwidth]{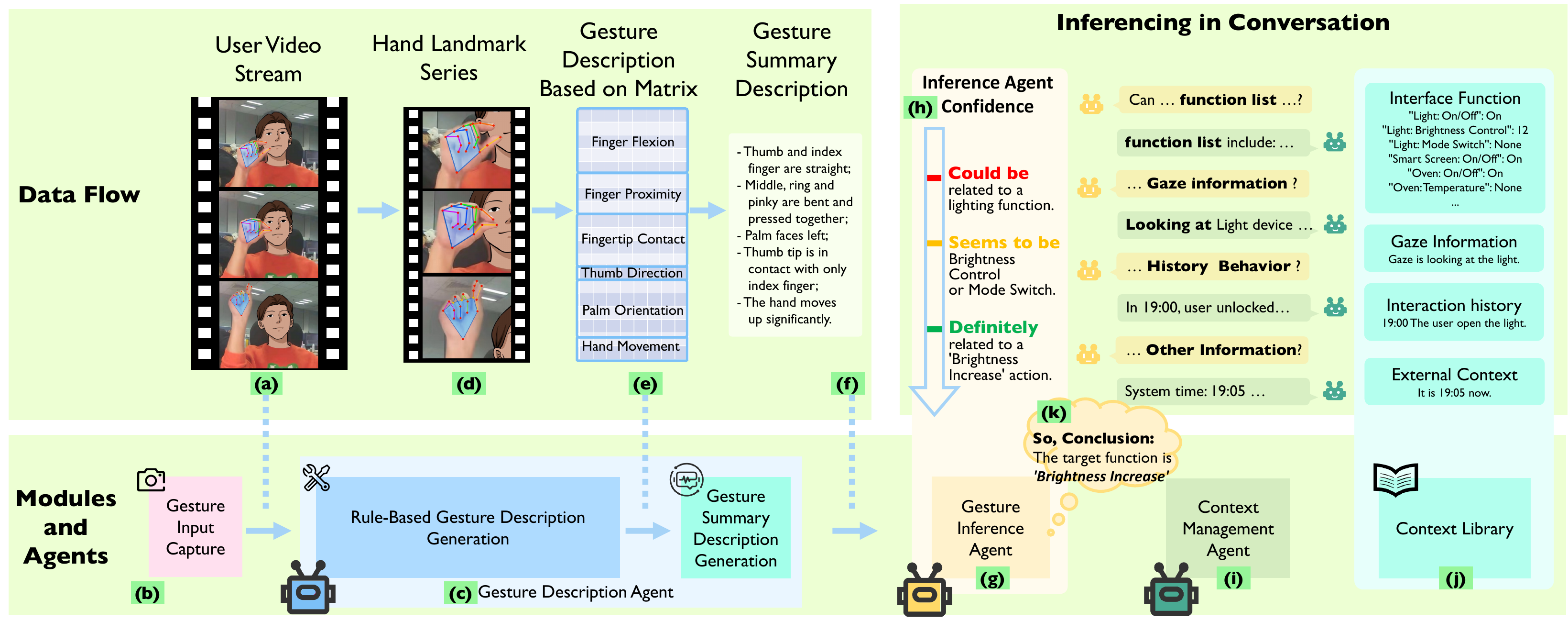}
  \caption{
  This scenario illustrates a system implemented based on GestureGPT. 
  }
  \label{fig:teaser}
\end{teaserfigure}

\maketitle

\section{Introduction}
\label{sec:introduction}
Gestures express human intent intuitively and promptly, enabling natural human-computer interaction with low cognitive load~\cite{pavlovic_visual_1997, xia_iteratively_2022}.
Most existing gesture interfaces support only predefined gestures, which can achieve a very high gesture classification accuracy~\cite{wah_ng_real-time_2002, berman_sensors_2012}.
However, it requires significant effort to learn and memorize different gestures and their respective mappings to system functions, particularly with a large set of gestures~\cite{gao_working_2023, ejaz_effect_2019}.
The gesture-function mapping is also fixed, which cannot be easily expanded and may contradict users' preferences~\cite{delamare_gesture_2019, vatavu_user-defined_2012, nacenta_memorability_2013}.

To address the limitations of predefined gestures, researchers have proposed gestural interfaces that support user-defined gestures~\cite{lu_gesture_2012, xu_enabling_2022, wang_gesturar_2021}.
Users can define their own gestures for each function with only a few demonstrations, thus eliminating the learning effort and enhancing system flexibility.
However, each user must design and demonstrate their own gesture and associate it with a system function~\cite{mo_gesture_2021, speicher_gesturewiz_2018}.
Users also still need to memorize their own gestures, which may lead to frustration using the interface, especially when there are a large number of functions.  
Such constraints degrade the natural interaction experience and hinder the broad adoption of gestural interfaces, leading to Norman's famous argument that ``Natural User Interfaces are Not Natural''~\cite{norman_natural_2010}.
While significant advances have been made in gesture recognition technologies over the years, this naturalness challenge of gestural interfaces persists.
To address the pursuit of more natural gestural expressions, users expect high-frequency gestures to seamlessly link various functions (e.g., common touchscreen gestures like zooming in/out, or typical semantic gestures like thumbs-up and rock-and-roll) without being confined to a limited set of expressions or frequently needing to customize their own~\cite{Jian-peng2019Research, Li2012Semi-customizable}. Additionally, as gestural interfaces become more widespread, users will likely experience significant improvements in both their familiarity with these expressions and the speed of their interactions~\cite{Stiegemeier2022User, Murray2010Freedom}.

To that end, future gestural interfaces require direct, end-to-end \textbf{gesture understanding} rather than merely \textbf{gesture recognition}.
Users should not have to learn, memorize, or demonstrate specific gestures.
Instead, they can perform gestures naturally, according to their understanding of the function's semantics and their everyday interaction experiences with humans and machines.
The interface automatically links the gesture to its corresponding function, considering both the gesture and the interaction context.
This gesture understanding task is inherently more challenging than simple gesture recognition.

We propose GestureGPT, an LLM-based paradigm that comprehends natural free-form hand gestures and automatically links them to their intended functions.
The core idea of GestureGPT is to utilize the LLM’s rich common sense for recognizing gestures and understanding the interaction context, as well as its potent inference capabilities to map gestures to their intended functions.
Constructing such an LLM-based gesture understanding framework demands considerable effort.
For instance, gestures and context information must be transformed and formatted in a way that LLMs can process.
Additionally, there needs to be a mechanism for robust and thorough synthesis of all relevant information without overlooking any details.

To that end, we design a triple-agent paradigm that generates gesture descriptions, manages the context, and infers the intended function. This involves a \emph{Gesture Description Agent}, a \emph{Gesture Inference Agent}, and a \emph{Context Management Agent}, all based on LLMs.
1) The \emph{Gesture Description Agent} describes hand gestures in natural languages to facilitate the powerful text-based LLMs' reading and understanding of gestures.
A set of specially designed and tuned rules transforms hand poses and movements into discrete states (\eg the bending state of each finger) based on visually extracted hand landmarks, triggered by raising the hand to the chest level.
The agent then synthesizes a general description of the gesture based on a matrix formed with the states. 
2) The \emph{Gesture Inference Agent} analyzes the description and initiates a conversation with the \emph{Context Management Agent} to inquire about the context information.
After multiple rounds of dialogue, it infers which interactive function the user intends to activate.
3) The \emph{Context Management Agent} answers questions from the \emph{Gesture Inference Agent} by referring to a context library, which includes various types of context information, such as gaze points and interaction history, \etc~ The context library is organized in JSON file format. 

Figure~\ref{fig:teaser} envisions a process in which a system, implemented through the GestureGPT concept, understands a user’s gesture. In this scenario, he (a) triggers the system with a raised right hand above the chest, followed by looking at the light and performing a ``pinch and move up'' gesture. The (b) input camera captures this action through real-time video streaming. (c) The  \textcolor{description}{Gesture Description Agent} processes the gesture by (d) filtering key frames from the video and extracting hand landmarks to (e) generate the gesture state matrix. Subsequently, (f) \textcolor{description}{Gesture Description Agent} generates a description of the gesture’s pose and movement. Upon activation by the description, the (g) \textcolor{inference}{Gesture Inference Agent}  analyzes this description to determine which function the gesture maps to and (h) assesses the confidence level of the result. If the confidence is deemed insufficient,  \textcolor{inference}{Gesture Inference Agent} requests context information from the (i) \textcolor{context}{Context Management Agent}, which retrieves relevant data from the (j) context library containing all available context information in the current environment, to inform the inquiry. According to the inquiry, as context information such as ``Gaze: user is looking at the light'' and ``History: the user turned on the light at 19:00'' is incorporated, the \textcolor{inference}{Gesture Inference Agent}  (k) concludes that the function designated by the gesture is to ``increase the brightness'' of the light. The function could then be triggered to finish the interaction process.

GestureGPT offers a number of advantages that address the challenges and limitations of previous gestural interfaces. 
1) GestureGPT allows users to perform natural free-form hand gestures, which do not need to resemble those in a predefined gesture set or those previously defined by users.
This eliminates the need for learning, memorizing, or demonstrating specific gestures.
2) GestureGPT automatically associates gestures with their corresponding interface functions through a step-by-step inference process based on both the gesture description and the interaction context.
Such a novel LLM-based structure successfully addresses the ephemeral nature of gestures and their close relationship with context~\cite{norman_natural_2010}. 
Additionally, the integration of context further enhances the accuracy of gesture mapping.  
3) GestureGPT analyzes spatial coordinates of hand landmarks to generate gesture descriptions, making it independent of view angles and even modalities. For example, hand landmarks can also be reconstructed by wearable sensors~\cite{jiang_stretchable_2020, zhang_findroidhr_2018}. Such flexibility makes it easy to adapt GestureGPT for a wide range of applications. The natural language descriptions also preserve user privacy and reduce data transmission load, which are important for interaction interfaces.

Describing nuanced finger states and movements is crucial for accurately understanding hand gestures.
Our hand landmarks-based method captures finger states accurately and thoroughly by first calculating a set of discrete finger states based on rules, then using an LLM to synthesize a gesture summary.
The rules' parameters are tuned using third-person view public gesture image datasets and tested on both third-person view (HaGRID~\cite{kapitanov_hagrid_2024}, 38576 test samples) and first-person view datasets (EgoGesture~\cite{zhang_egogesture_2018}, approximately 5000 test samples).
The error rates are 2.3\% for third-person views and 6.3\% for first-person views, respectively.
Two gesture experts rated the synthesized summary, which achieved a score of 3.51 (\(\text{std} = 1.14\)) on a 5-point Likert scale, where 1 indicates that almost none of the description is correct and relevant, and 5 indicates that almost all of it is.

To evaluate our framework, we conducted experiments under two realistic interaction scenarios: smart home IoT device control using an AR headset (first-person view with 18 functions) and online video streaming on a desktop PC (third-person view with 66 functions).
The highest zero-shot Top-1/Top-5 gesture grounding accuracies achieved were 44.79\%/83.59\% for smart home control and 37.50\%/73.44\% for video streaming.
We report accuracies at various levels to more comprehensively understand and demonstrate the framework's performance and boundaries.
The results demonstrate the significant potential of GestureGPT, which, to the best of our knowledge, is \textbf{the first zero-shot free-form gesture understanding framework} that requires no learning, memorization, demonstration, or association efforts.

However, GestureGPT is not yet a practical interface ready for everyday use, primarily due to the slow inference speeds of current LLM systems.
Our current system averages 227 seconds per task in our evaluations, a delay caused by long prompts, multi-turn dialogues, and rate limits enforced by LLM cloud services.
Instead, it introduces a new \textbf{paradigm} for gestural interfaces.
To the best of our knowledge, GestureGPT is the first feasible framework for the inherently difficult task of understanding free-form gestures.
Previous efforts in gesture recognition have not successfully addressed free-form gestures, let alone the more complex challenge of gesture understanding.
GestureGPT, therefore, not only pioneers a new approach but also lays the foundational framework for future advancements. This framework can also be easily expanded to recognize a broader range of user intents through different modalities, such as facial expressions and body postures, among others.
Technologies such as edge-side LLMs or specifically fine-tuned large multi-modal models could significantly reduce response times, paving the way for a more practical implementation of this paradigm in the future.

Our contribution is three-fold:
\begin{enumerate}
    \item We proposed and evaluated the first framework that mimics human reasoning processes to achieve automatic free-form hand gesture understanding, as to our best knowledge. 
    \item We designed a set of gesture description rules based on hand landmarks to thoroughly and accurately capture states and movements of the hand, which has comparable performance to SOTA large multi-modal modal GPT-4o. 
    \item We carefully crafted prompts for each agent, analyzed the offline evaluation results, and generated various insights. Such insights are invaluable for future context-aware agent-based gesture understanding work. 
\end{enumerate}

\section{Background and Related Work}

\subsection{LLM as Autonomous Agent} 
Large language models have displayed an exceptional ability to understand and execute a broad spectrum of tasks~\cite{openai_gpt-4_2024,bai_constitutional_2022}. LLM has the potential to emulate human-level intelligence, accurately perceive generalized environments, act accordingly, and iterate to enhance outcome~\cite{shinn_reflexion_2023} when faced with diverse situations~\cite{wang_survey_2024}. 
Agents based on LLMs have shown potential in various domains, from web browsing~\cite{yao_webshop_2022,deng_mind2web_2023, zheng_gpt-4vision_2024}, strategic planning~\cite{yao_react_2023} to robotic control~\cite{driess_palm-e_2023,brohan_rt-2_2023}. 
This paper, on the other hand, addresses a notable gap in existing literature and utilizes LLMs for free-form gesture understanding and interaction. 

Though LLM agents have shown promising intelligence, a single agent implementation may suffer from performance degradation in long context scenarios~\cite{jiang_longllmlingua_2023}. 
Instead, multi-agent systems have shown superiority to accomplish more complex tasks in collaboration, reducing hallucination~\cite{wang_unleashing_2023}, and information exchange~\cite{talebirad_multi-agent_2023}. 
For instance, \citet{park_generative_2023} designed a multi-agent system that simulates human behavior in a virtual environment. 
\citet{park_choicemates_2023} relies on conversational interactions among multiple agents to aid online decision-making, which shows multi-agent's great potential in reasoning under unfamiliar scenarios. 
Other applications software development~\cite{qian_communicative_2023}, span reasoning~\cite{liang_encouraging_2023}, evaluation~\cite{chan_chateval_2023,zhang_wider_2023}, and a myriad of intricate tasks~\cite{zhuge_mindstorms_2023, park_choicemates_2023}.
So, GestureGPT chooses a triple-agent architecture to better handle the complicated context-aware gesture understanding task. 
The goals of the three agents are clearly defined and isolated, so that they can be optimized individually while collaborate seamlessly to achieve accurate gesture comprehension.

\subsection{Natural Free-form Gesture Understanding}
\label{sec:relatedword_zeroshot}

Gestural interfaces working with pre-defined gesture set demand large annotated dataset, and the confinement to pre-defined gestures also hampers the naturalness of interaction~\cite{wexelblat_approach_1995}. 
The conflicts in different system designs further exacerbate user adaptation challenges across platforms.
User-defined gestural interfaces mitigates the learning burden by allowing users to define their own gestures.
Only several demonstrations of the gesture are necessary for the system to learn new gestures with the help of advanced few-shot learning algorithms~\cite{xu_enabling_2022}. 
Gesture Coder~\cite{lu_gesture_2012} allows the user to demonstrate a gesture, and, instead of defining a gesture name for it, the user directly associate it to a designated function with auto-generated recognizer.
But in these works, users still need to demonstrate and memorize the gesture while ensuring its distinctiveness from existing gesture commands. 

Nevertheless, all aforementioned approaches are still limited by a finite set of distinguishable gestures defined by interface designers or users.
This leads to a mismatch between the fixed gesture set and the flexible gestures humans perform in different scenarios, greatly restricting gesture expressiveness.
Thus, both pre-defined and user-defined gestural interfaces require users to adapt to them, rather than the reverse~\cite{wexelblat_approach_1995}.
Free-form gestural interfaces, however, do not have these limitations.
For instance, Gesture Avatar~\cite{lu_gesture_2011} enables any drawing gesture on a screen.
The input gesture is linked to a UI element based on appearance resemblance, providing an intuitive interaction experience.
Yet, gaps remain in the research of free-form hand gesture understanding.
Compared with screen gestures, hand gestures are more complex, do not necessarily resemble UI elements, and can vary under different scenarios for the same function.

Recent advances in zero-shot learning make it possible to recognize unseen gesture classes~\cite{madapana_database_2019, madapana_hard_2018, wu_prototype-based_2021}. 
Madapana~\cite{madapana_database_2019, madapana_hard_2018} proposed frameworks that define a set of attributes for gestures, recognizing unseen gestures by manually labeling these attributes for each unseen gesture.
Similarly, \citet{wu_prototype-based_2021} introduced a prototype-based approach that requires the definition of ``semantic prototypes'' for unseen gesture categories. 
These methods explore to some extent the recognition of ``unseen'' gestures, enabling the system to recognize new gestures without additional training.

However, above methods focus solely on the task of gesture recognition. Even for the recognition task, they only identify key features of new gestures, not their semantic names. 
Gesture understanding, however, requires not only recognizing gestures but also mapping them to functions, which demands a comprehensive integration of complex contextual information (such as available functions, interaction history, and physical environment). 
Such analysis and reasoning necessitate mimicking human thought processes, a task that traditional machine learning models struggle to achieve.

\begin{figure}[!h]
\vspace{-5pt}
  \centering
  \includegraphics[width=0.6\textwidth]{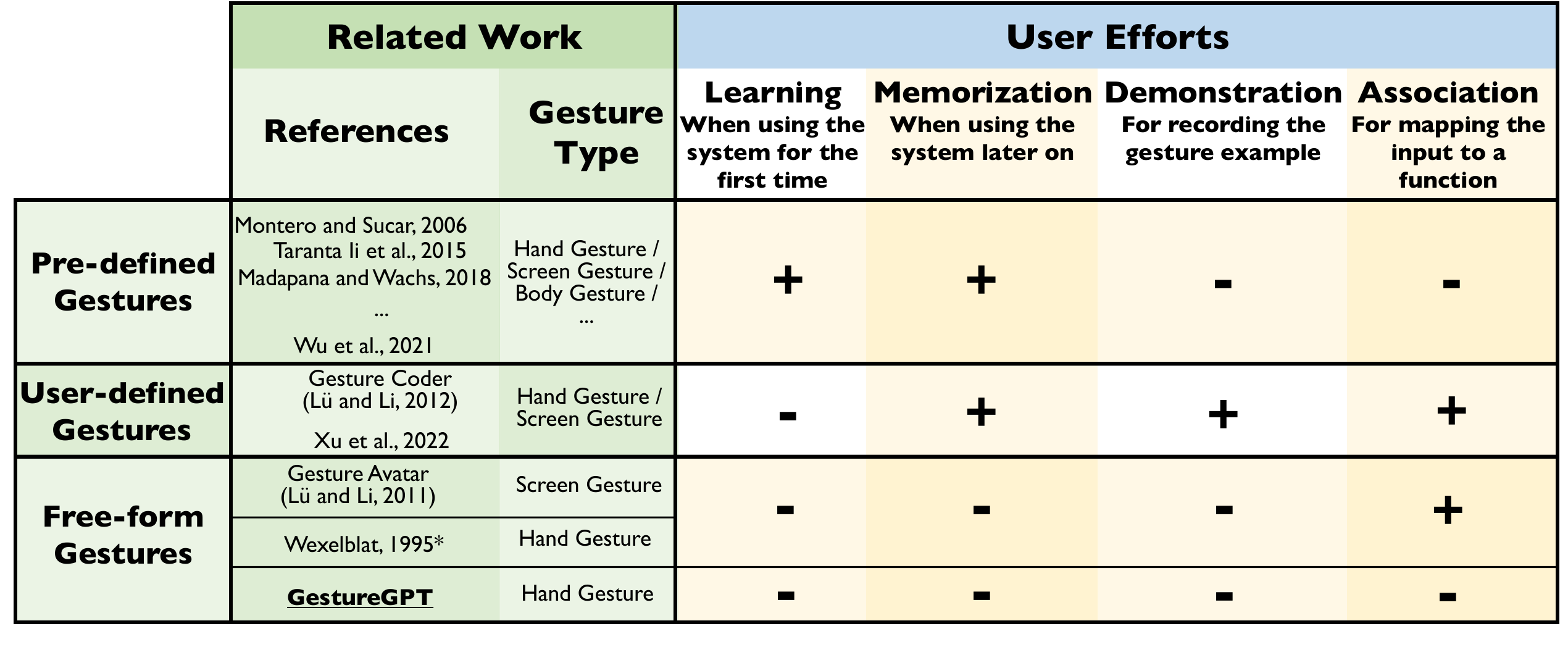}
  \caption{Overview of current gestural interaction systems.}
  \label{fig:related_work_graph}
\vspace{-15pt}
\end{figure}

To tackle such challenges, our framework relies on LLMs to understand the semantic meanings of both the hand gestures and the system functions to ensure correct gesture-function mapping, because LLMs are currently the most promising tools capable of human-alike inference and comprehensive context understanding.
An overview of current gesture-based interaction systems and the placement of our work is shown in Figure \ref{fig:related_work_graph}.

\subsection{Gesture Understanding with Context Information}
As Norman pointed out, gestures are ephemeral in nature and highly context related~\cite{norman_natural_2010}. 
They inherently possess diverse semantics across different contexts, and may embody social metaphors~\cite{sun_metaphoraction_2022}. 
Contextual information such as spatial distance~\cite{hutchison_context-based_2006}, gaze~\cite{chatterjee_gazegesture_2015}, speech~\cite{wexelblat_approach_1995}, user history~\cite{morency_head_2006, chen_gap_2023} and domain-specific knowledge~\cite{taranta_ii_exploring_2015} have been integrated into gestural interaction systems to improve gesture recognition accuracy. 
However, most previous work only leverage gaze information for simplified tasks~\cite{schwarz_combining_2014, ge_improving_2023, huang_using_2015, akkil_gaze_2016, singh_combining_2018}.  
The handling of different types of context typically require highly specialized models~\cite{singh_combining_2018, idan_network-based_2022, chen_gap_2023}, which limits the generalizability of such systems. 

LLM agents' promising ability to solve complex problems provides an alternative solution for inference based on different types of context information. 
LLM agents can retrieve higher-level context like semantic meaning of interaction elements~\cite{gur_understanding_2023} and users’ profile and preferences~\cite{rivkin_sage_2024}. 
The way it utilizes context and the intentions it can deal with are also greatly enlarged. 
For example, \cite{king_sasha_2024, rivkin_sage_2024, gur_real-world_2024, king_get_2023} have shown that LLM agents can turn loosely-constrained commands into appropriate actions. 
Inspired by existing research, we use a \emph{Context Management Agent} to manage context information stored in the context library, and a \emph{Gesture Inference Agent} to infer the intention behind a gesture based on context cues.

\section{Method}

\begin{figure*}[!h]
 \centering
 \includegraphics[width=\textwidth]{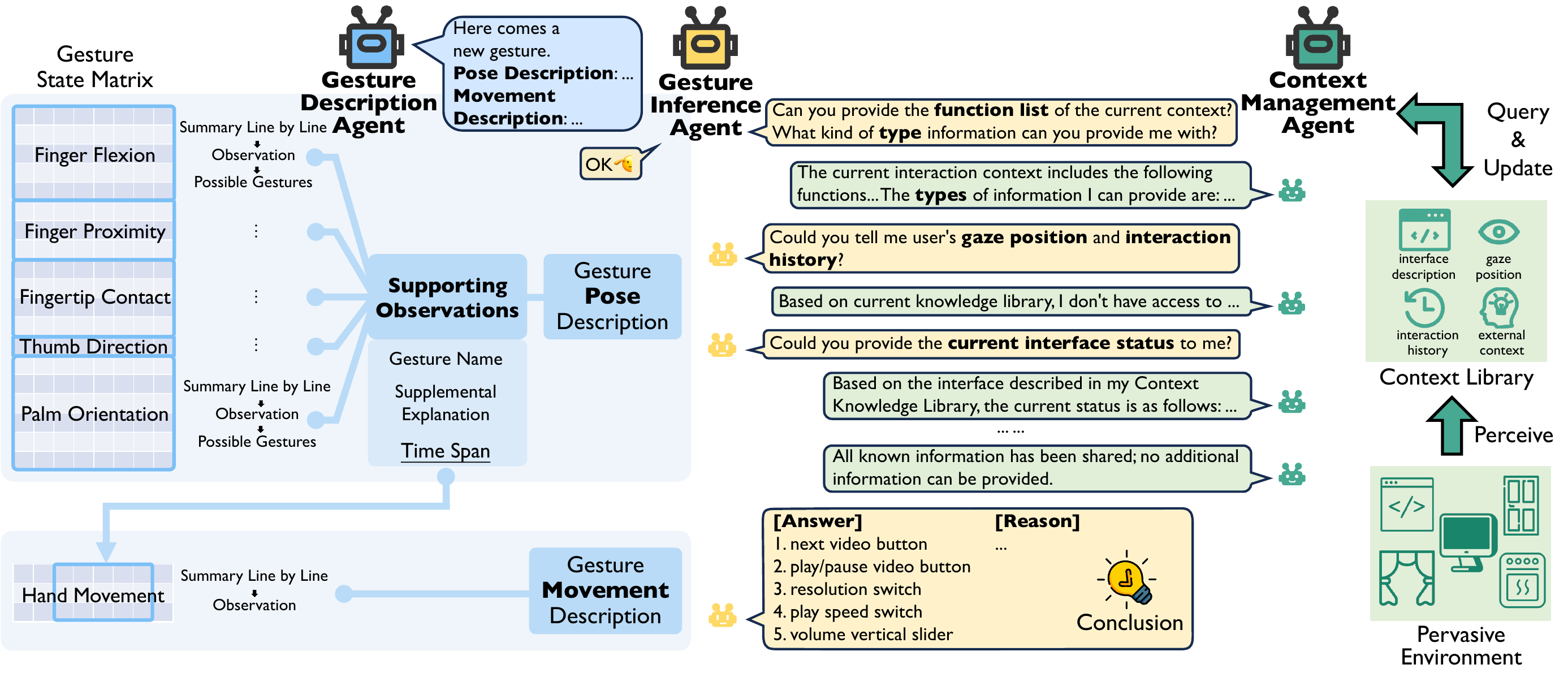} 
 \caption{Agents Collaboration Workflow.}
 \label{fig:system_arch}
\end{figure*}

GestureGPT has a triple-agent framework to efficiently manage gesture description, context, and inference tasks, enhancing system flexibility and scalability while addressing the shortcomings of single or dual agent systems, consists of: 
(1) \textbf{\textit{Gesture Description Agent}}, generating descriptions of gestures from videos; 
(2) \textbf{\textit{Context Management Agent}}, handling interaction context; 
and (3) \textbf{\textit{Gesture Inference Agent}}, synthesize information to infer gesture intentions through dialogue and reasoning.

The workflow initiates with the transformation of user gestures, captured by an RGB camera, into natural language description. 
The \textit{Gesture Description Agent} first uses a set of rules to transform the gesture into a ``Gesture State Matrix'', which delineates hand and finger states over time.
The agent then relies on this matrix to produce a summary of gestures, which is forwarded to the \textit{Gesture Inference Agent}.
Upon receiving the gesture description, the \textit{Gesture Inference Agent} engages in multi-round dialogue with the \textit{Context Management Agent}.
This involves identifying and soliciting relevant context information from a context library managed by the \textit{Context Management Agent}.
Through iterative dialogue, the \textit{Gesture Inference Agent} gains a comprehensive understanding of the interactive scenario and completes a dynamic mapping between the gesture and possible functions.

\subsection{Gesture Description Agent}

The Gesture Description Agent is crucial for translating video-captured gestures into natural language descriptions that is understandable by LLMs. 
We choose LLM since it has richer common knowledge and stronger inference than existing Large Multimodal Models (LMMs) that we can access, which is vital to deal with context-aware free-form gesture understanding tasks.

\subsubsection{Rule-Based Gesture Description Generation}

We design a set of 6 rules to encode the hand gesture in terms of finger flexion, proximity, contact, direction, as well as palm orientation and hand position (Table~\ref{table:gesture_rule}). The detailed rule definition can be found in Appendix~\ref{apx:rule_calculation_description}.

\begin{figure}[htbp]
  \vspace{-10pt}
  \centering
  \includegraphics[width=0.8\textwidth]{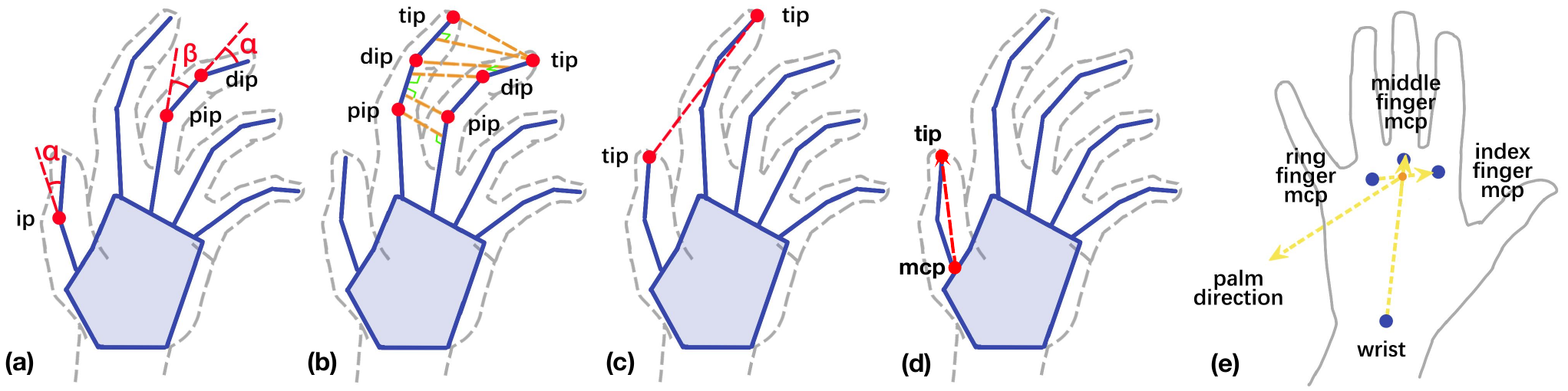}
  \caption{Illustration of gesture description rules. (a) The flexion of a finger is calculated from the sum of bending angle of ip joint (for thumb) or pip and dip joint (for other fingers). (b) The proximity of two fingers is calculated from the average distance from each finger's pip/dip/tip joint to the other finger. (c) The contact of thumb and another finger is calculated from the distance of their fingertips. (d) The pointing direction of thumb is calculated from the vector from thumb's mcp to tip. (e) The palm orientation is calculated from the dot product of the two vectors on the hand, pointing towards the reader. }
  \label{fig:rule_calculation}
  \vspace{-10pt}
\end{figure}

Each state is calculated using the coordinates of hand landmarks, as shown in Fig~\ref{fig:rule_calculation}, which are generated by MediaPipe~\cite{zhang_mediapipe_2020} from videos captured with an RGB camera.
For each rule, parameters (such as the distance between fingers to determine their proximity) are trained on a third-person view (\ie viewed from an external perspective, typically through a camera not aligned with the subject’s viewpoint) public gesture dataset, and tested on first-person view (\ie viewed from the subject's own perspective, typically through a head-mounted camera) and third-person view datasets. 
Test results show an error rate of 2.3\% for third-person view and 6.3\% for first-person view samples, indicating the rules' efficacy in retaining accurate information from gestures across different viewpoints, validating it a universal solution applicable to various gesture capture devices. 
The details of parameter training and the evaluation metrics for these rules are thoroughly described in Appendix~\ref{apx:rule}. The pseudocode for the rules is provided in Appendix~\ref{apx:pseudocode-rules}. An evaluation and error analysis of the rules can be found in Appendix~\ref{apx:rule_training_and_evaluation}.

\begin{table}[!h]
  \centering
  \caption{Summary of rules, their meanings, and corresponding values.}
  \label{table:gesture_rule}
  \begin{tabularx}{\columnwidth}{|X|X|X|X|}
  \hline
  \textbf{Rule Name} & \textbf{Applicable to} & \textbf{Calculation} & \textbf{Value} \\
  \hline
  Finger Flexion  & Thumb, Index, Middle, Ring, Pinky & measured as the total bending angle of each joint: for the thumb, the IP joint; for other fingers, the PIP and DIP joints. & 1: Straight; \newline 0: Between; \newline -1: Bent \\
  \hline
  Finger Proximity  & Index-Middle, Middle-Ring, Ring-Pinky & calculated as the average minimal distance from each finger’s joint to the other finger. &  1: Pressed Together; \newline 0: Between; \newline -1: Apart \\
  \hline
  Thumb Fingertip Contact  & Thumb-Index, Thumb-Middle, Thumb-Ring, Thumb-Pinky & computed as the distance between their fingertips. & 1: Contact; \newline 0: Between; \newline -1: No Contact \\
  \hline
  Pointing Direction & Thumb &  the direction from thumb’s mcp joint to tip joint. & 1: Upward; \newline -1: Downward; \newline 0: Other Directions/Bent \\
  \hline
  Palm Orientation & Palm &  computed as the direction to which the palm is facing.  & One-hot encoding: [Left, Right, Down, Up, Inward, Outward]; \newline All zeros: Unknown \\
  \hline
  Hand Position & Hand & computed as the geometrical center of a hand by taking average of all 21 landmarks’ coordinates. & Float coordinates \\
  \hline
  \end{tabularx}
\end{table}

The gesture period of interest starts when users raises their hands to or above their chest level. 
Frames within a gesture period are processed at 0.2-second intervals, with each sampled frame undergoing rule-based calculations. 
However, the natural language descriptions concatenated from each rule are too long to fit into a prompt of an LLM. 
So we introduce a gesture state matrix that contains vectors of each frame, which capitalizes on LLMs' proficiency with code-formatted content~\cite{gao_pal_2023}. The detailed definition of the gesture state matrix is given in Appendix~\ref{apx:rule_matrix}.

\subsubsection{Gesture Summary Description Generation}
The system needs to extract the rich information embedded in the matrix and generate a concise summary of gesture descriptions for the \textit{Gesture Inference Agent}. 
A significant challenge is the gesture state matrix data type; the pre-trained data of the LLM likely does not contain exactly similar data. 
We employ a chain-of-thought~\cite{wei_chain--thought_2023} process to guide the LLM step-by-step, supplemented by high-quality expert examples that demonstrate how the analysis should be conducted. 
We choose to use two prompts to generate the hand \textbf{pose} (finger flexion, finger proximity, fingertip contact, thumb direction and palm orientation) and hand \textbf{movement} description separately. 
As is shown in Figure~\ref{fig:system_arch}, the Gesture Description Agent first converts pose-related rows of the matrix to descriptions of possible gestures, along with the time span during which this gesture happened. 
The time span is then used to select the corresponding part of movement-related rows, and another prompt is used to convert it to movement description. 
As long as the agent correctly finds the valid pose of the gesture, irrelevant movements are naturally filtered out.

\begin{figure}[htbp]
\centering
    \begin{minipage}[b]{0.42\textwidth}
      \centering
      \includegraphics[width=\textwidth]{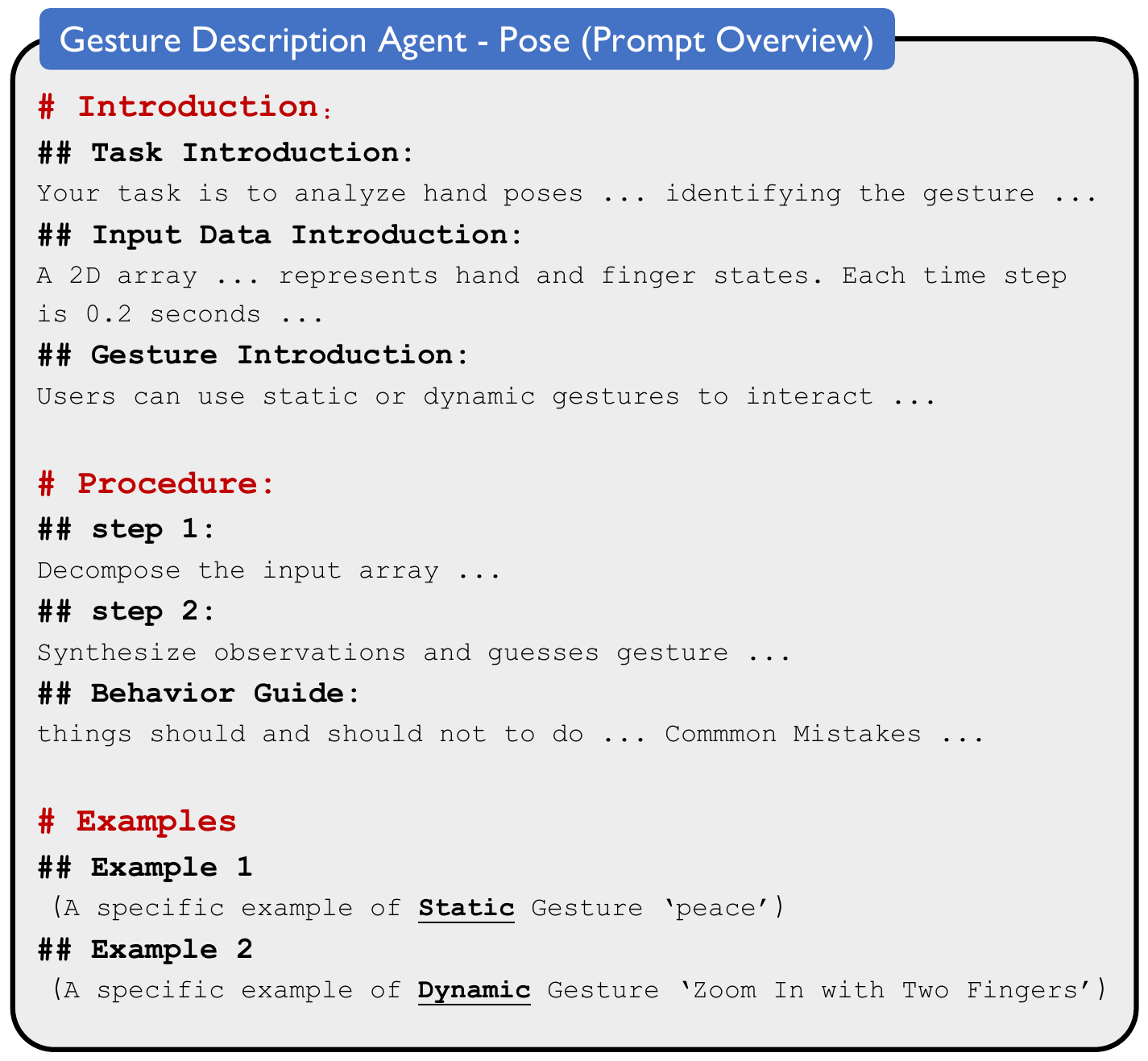}
      \caption{Prompt for Gesture Description Agent.}
      \label{fig:prompt_description_agent}
    \end{minipage}
    \begin{minipage}[b]{0.42\textwidth}
      \centering
      \includegraphics[width=\textwidth]{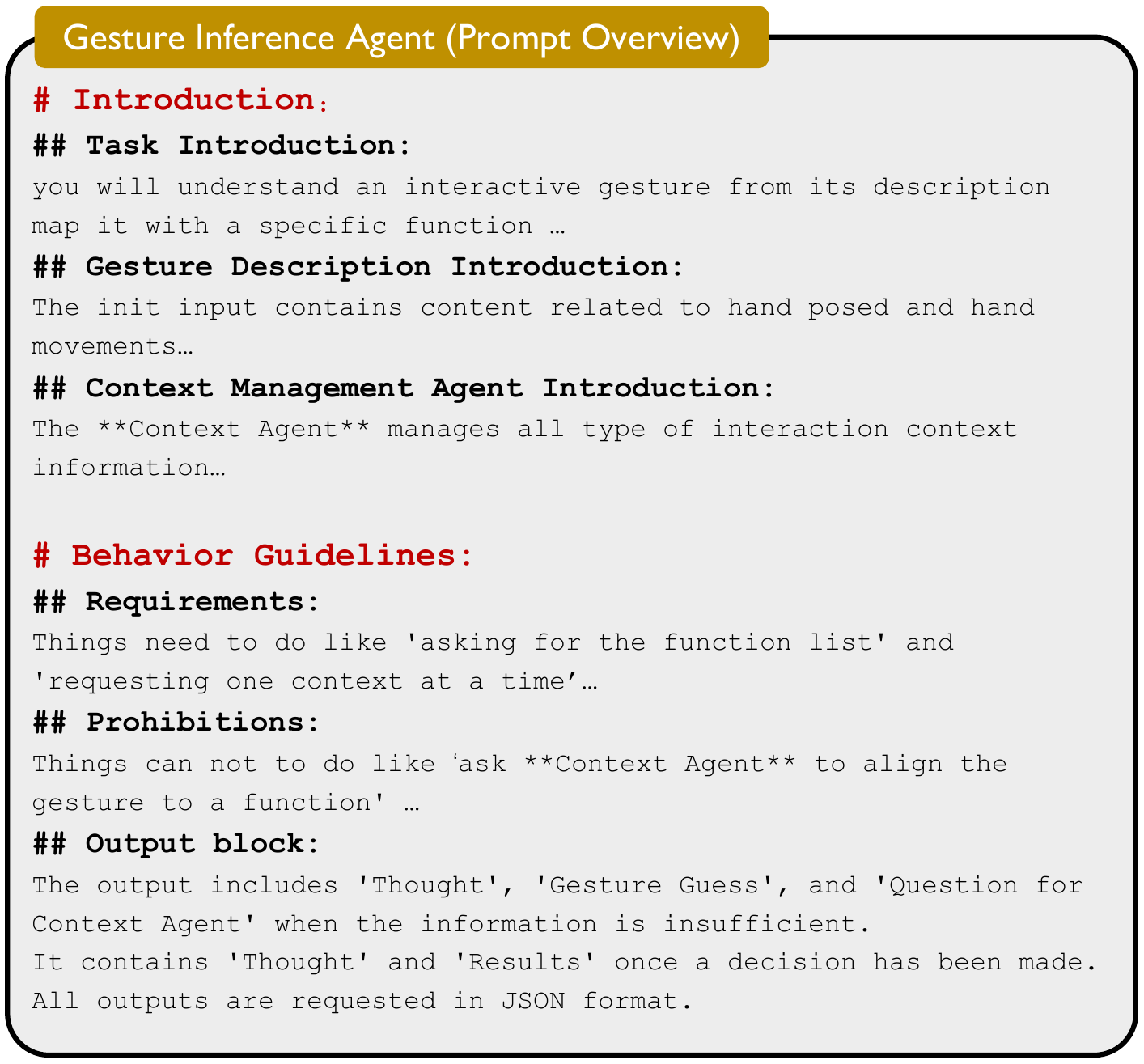}
      \caption{Prompt for Gesture Inference Agent.}
      \label{fig:prompt_inference_agent}
    \end{minipage}
\vspace{-10pt}
\end{figure}

For pose description, we developed a prompt structured into three parts shown in Figure~\ref{fig:prompt_description_agent}: Introduction, Procedure and Examples.
The introduction outlines the task, describes the the gesture state matrix, and explains the concept of ``interactive gestures'' versus non-ineractive hand movements, helping the agent to distinguish between different types of gesutres and understand their lifecycle. 
The procedure section teaches the agent to understand the gesture state matrix and guess possible gestures from the matrix, while addressing some common mistakes to regulate its behaviors. In paticular, it instructs the agent to decompose the matrix, guess possible gestures from each part, and synthesize the conclusions from all parts to get the most possible gestures, encouraging a human-like process of understanding and summarizing data, while using transcription to counter LLM's forgetfulness and improve accuracy. 
The last part provides two typical examples of static and dynamic gestures to enhance LLM performance by reflecting back to the guidelines stated before.
This well-structured prompt ensures that the LLM model correctly processes the matrix data type. 
The hand movement description prompt shares a similar structure. The generated description of the gesture will be like: 

\begin{minipage}{\linewidth}
\small{
\begin{verbatim}
- Thumb and Index Finger transition from non-specific/bent to straight.
- Middle, Ring, and Pinky Fingers transition from bent to straight.
- Fingers start close together and then spread apart.
- Thumb starts in contact with all fingertips but then moves away.
- Palm orientation starts facing left but becomes non-specific.
- The hand moves left slightly with negligible vertical and depth movements.
\end{verbatim}
}
\end{minipage}

\subsection{Gesture Inference Agent}

Upon receiving gesture descriptions from Gesture Description Agent, the Gesture Inference Agent engages in a dialogue with Context Management Agent.
This conversational exchange is pivotal for the Gesture Inference Agent to discern the user's actual intent in the context, \ie associating the gesture with a target system function from a  list of potential functions supplied by Context Management Agent. 
This setup epitomizes a collaborative conversation aimed at overcoming the challenges of gesture ambiguities under different interactive scenarios and context. 

We outline the agent's tasks and some behavioral guidelines to better demonstrate the semantic nature of the agent task revolves around thinking, summarizing, and inferring. 
The overall prompt is structured into two parts shown in Figure~\ref{fig:prompt_inference_agent}.

The introduction part includes a Task Introduction, a Gesture Description Introduction, and a Context Management Agent Introduction (its conversational counterpart). 
A critical aspect to convey in this part is that the initial gesture description may contain errors. Therefore, the Gesture Inference Agent must utilize context information to help with the inference, correct potential mistakes, and make decisions.
Behavioral guidelines for the agent consist of seven directives and five prohibitions, acquired by iteratively updating the prompts during implementation. When new error tendencies are observed, corrections are introduced into the prompt, akin to a teacher's guidance. 
Then, we define the output format. 
We opt for JSON due to its ease of parsing, allowing for efficient extraction of useful information for subsequent conversational rounds. Importantly, we require the agent to first articulate its \textbf{thoughts} before posing \textbf{questions} or drawing \textbf{conclusions}. This approach has been proven to make the agent's behavior more reasonable.
Once the agent deems that it is confident to decide or that no further context can be gleaned, it proceeds to make a final decision, listing the top-5 possible functions for the current interface, ranked from most to least likely.
The Gesture Inference Agent emphasizes the importance of precise communication, contextual reasoning, and analytical capabilities in interpreting and responding to gesture-based interactions.

\subsection{Context Management Agent}

The Context Management Agent plays a crucial role in our system, offering intelligent management of various types of context information. 
Unlike static, rule-based context management system, this agent dynamically adapts to unfamiliar context thanks to LLM's vast and intricate knowledge base, ensuring a flexible and responsive interaction environment.

\begin{figure}[htbp]
  \vspace{-10pt}
  \centering
  \includegraphics[width=0.35\textwidth]{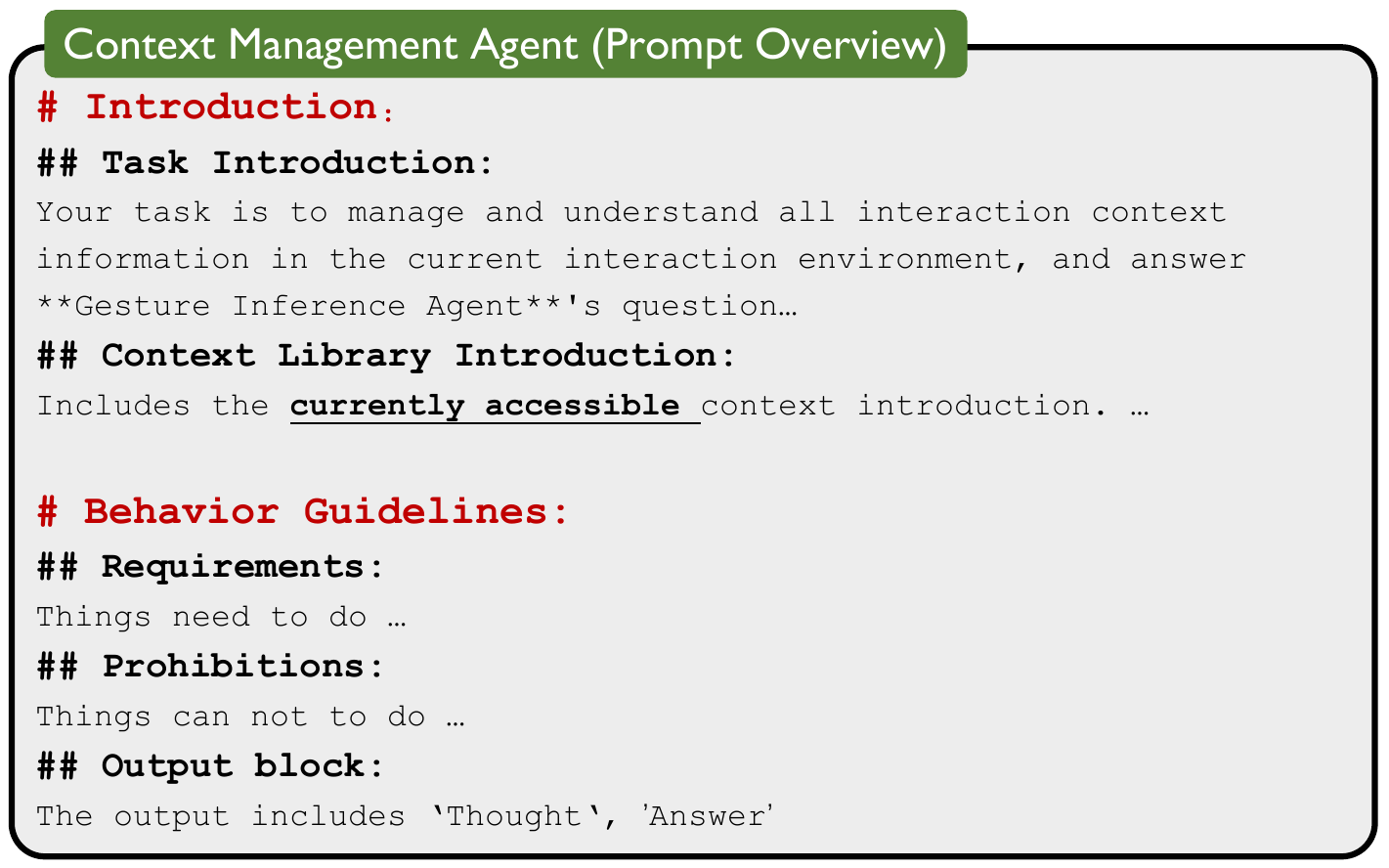}
  \caption{Prompt for Context Management Agent.}
  \label{fig:prompt_context_agent}
  \vspace{-10pt}
\end{figure}

A dedicated Context Management Agent represents a significant step forward in the evolution of gesture-based interfaces. 
By segregating context handling from the inference processes, we streamline system operations and enhance the overall system's ability to effectively utilize diverse contextual data. 
This separation allows for more sophisticated and nuanced interaction paradigms, where gestures can be interpreted in varying scenarios with greater precision. 
Moreover, this architecture facilitates easier updates and scalability, as new contexts or rules can be integrated without disrupting the core inference mechanisms. 
Consequently, this approach not only addresses previous limitations but also sets a new standard for the development of adaptive, efficient, and user-centric gesture recognition systems. The overall prompt is structured into two parts shown in Figure~\ref{fig:prompt_context_agent}.

The introduction includes a Task Introduction and a Context Library Introduction. The latter introduces the currently accessible contexts and is designed to be adjustable.
Similar to the Gesture Inference Agent, iterative implementation guides the Context Management Agent's requirements and prohibitions, with 4 requirements and 2 prohibitions.
The output is structured in JSON format for easy parsing, allowing efficient information extraction and facilitating subsequent conversation rounds.

\subsubsection{Context Library Operations}
\label{section:method_context_operations}
The Context Management Agent supports three main operations within the context library: \textbf{Adding} new context types, \textbf{retrieving} context information based on queries and \textbf{calculating} specific context values.

\begin{itemize}
    \item \textbf{Adding Context Types:} Context types and their values are organized using markdown for text and JSON for structured data. New context types are introduced with natural language descriptions, and associated values are formatted in JSON. This facilitates easy system expansion and automatic incorporation of new contexts into the operational prompt.
    
    \item \textbf{Retrieving Context Information:} The agent automates retrieval operations, streamlining the process by organizing data in a standardized format (\eg JSON), which the LLM can autonomously interpret.

    \item \textbf{Calculating Context Value:} The agent can calculate specific context values based on predefined methods in the context's description. These methods can be implemented in a Python script. Upon activation of a calculation operation, the agent generates a unique placeholder. This placeholder triggers the system to execute the Python script, automating the retrieval of context values and producing the desired output. The placeholder is then replaced with this output in the final response.
\end{itemize}

The design of Context Management Agent offers significant benefits, such as the ease of adding new context types using simple natural language descriptions and the automated retrieval and understanding of context information. This enhances the system's versatility and usability. 
Our design also inform the design of context management components of future interactive systems.

\section{Evaluation}

We designed two experiments to assess GestureGPT's adaptability and effectiveness under different interaction scenarios with varying context environments and camera perspectives.

In the first interaction scenario, a user controls smart home appliances through an AR headset. 
In this setup, gestures are captured from the user's own viewpoint, offering a first-person perspective. 
It is a key interaction scenario in the future when users interact with environment using head-mounted devices. 

Our second interaction scenario mimics the case when a user is watching online videos. 
The user watches the video on a monitor with a camera capturing gestures from the third-person perspective. 
This setup is prevalent in a variety of settings, including smart TVs, interactive public displays, educational environments, gaming, \etc
~The complexity increases in this scenario, as there are plenty of functions and interactive elements on a webpage. 

We selected four contexts in our experiments to evaluate GestureGPT's performance in different context settings:
\begin{itemize}
    \item \textbf{Interface Function List:} Crucial for mapping gestures to interface functions, this context includes interface name and a list of functions with their names, locations, and unique IDs, key for navigating the user's interaction environment.
    
    \item \textbf{Gaze Information:} Data on the user's gaze, given in 3D (in home scenario) or 2D (in video scenario) coordinates.
    
    \item \textbf{Interaction History:} A record of the user's recent interactions.
    
    \item \textbf{External Context Information:} Information from other devices or sensors. We introduce this type of context to explore how other factors impact gesture understanding and whether the agent can leverage this information.
\end{itemize}

We informed participants that both video and eye movement data would be collected during the study, and we assured them of the confidentiality and safety of their data. The study is IRB approved by our local institution. All participants provided informed consent. 
In both experiments, we asked participants to perform the gesture as they see fit to finish the task. Both static hand poses and dynamic gestures involving hand movements were permitted. 

We employed the OpenAI GPT-4 model as the underlying architecture for our triple-agent system. Specifically, we utilized the \texttt{gpt-4-1106-preview} version for our evaluations. 
We configured the request temperature to 0 so that the model's output is as consistent as possible. 
Apart from this, we adhered to OpenAI's default settings for other parameters. 
To mitigate the effects of randomness and enhance the reliability of our findings, we ran the experiment repeatedly for three times. The aggregated results from these iterations were used to substantiate our conclusions.

\subsection{Experiment 1: Augmented Reality-Based Smart Home IoT Control}

This study explores the interaction between users and smart home devices through augmented reality (AR), specifically using gesture controls in a simulated kitchen environment. 
Participants perform gestures as they see fit to control various IoT devices, triggering changes in device state accordingly.

\subsubsection{Experiment Setting and Procedure}

\begin{itemize}
  \item \textbf{Experimental Platform and Data Collection} - The Microsoft HoloLens 2 served as the primary experimental platform, offering APIs to capture user gaze and hand gesture data accurately. The experimental environment was developed using Unity (version 2020.3.24f1), the Mixed Reality Toolkit (MRTK 2.8.0), and the OpenXR Plugin (1.7.0). The devices were represented by 3D models anchored in the space: a light, a smart cabinet, a smart screen, an oven, and an air cleaner.

  \item \textbf{Context Library Setup} - The experiment implemented context library as follows:
  \begin{itemize}
    \item \textit{Interface Function List}: Drawing from the XiaoMi SmartHome API\footnote{\url{https://iot.mi.com/new/doc/design/spec/xiaoai}}, five devices and their corresponding functions were synthesized to form a function list. Each device has 3-5 functions with a total of 18 functions in this scenario. 
    \item \textit{Gaze Information}: User gaze data was captured using the HoloLens 2 Gaze API and saved as 3D spatial coordinates. 
    \item \textit{Interaction History}: Interaction history was extracted from the task sequence.
    \item \textit{External Context Information}: There might be context information that is external to our system, which can significantly impact the grounding reasoning. We defined several external contexts corresponding to different tasks to understand if our system can correctly leverage those. 
  \end{itemize}

  \item \textbf{Task Descriptions} - Eight tasks were designed to simulate smart device control (Table~\ref{tab:smart_home_info}). 

  \item \textbf{Participants} - We recruited 16 participants from three local schools, compensating them at a rate of \$12 per hour. Their ages ranged from \(15\) to \(35\) years (\(\text{MEAN} = 26.625, \text{SD} = 5.325\)), comprising \(13\) males and \(3\) females. Participants included 1 high school student, 4 undergraduate students, 9 graduate students, and 2 research engineers. None of the participants had prior experience with AR/VR devices, which helped minimize bias due to varying familiarity with such technology.
  
  \item \textbf{Task Procedure} - Upon their arrival, participants were briefed about the scenario and the devices involved. They are asked to make any gesture deemed most intuitive using the right hand after raising their hand above their chest to initiate the trigger. A preliminary warm-up session was conducted to familiarize the participants with the AR devices and gesture control operations. Following this, they were instructed to complete the eight tasks. Feedback from the devices was simulated to enhance the interaction experience and realism of the study.

\end{itemize}

Detailed information on the experiment simulation interface, the list of devices and their functions, and the task list are provided in Appendix~\ref{appendix:exp_setting_home}. A total of \(16 \text{ participants} \times 8 \text{ tasks} = 128\ \text{gestures}\) were collected.

\subsubsection{Results Analysis}

\begin{table*}[h]
  \caption{Main Results of GestureGPT in the Two Experiments}
  \label{tab:main_result}
  \resizebox{\columnwidth}{!}{
  \begin{tabular}{lrrrr|rrrr}
  \toprule
   \midrule
  {} & \multicolumn{4}{c}{\textbf{Smart Home} Scenario} & \multicolumn{4}{c}{\textbf{Video Streaming} Scenario}\\
   \midrule
  {} & Top 1 ($\uparrow$) & Top 3 ($\uparrow$) & Top 5 ($\uparrow$) & Negative ($\downarrow$) & Top 1 ($\uparrow$) & Top 3 ($\uparrow$) & Top 5 ($\uparrow$) & Negative ($\downarrow$) \\
  \midrule
Random Guess&5.56\%&16.67\%&27.78\%&72.22\% &3.15\%&9.46\%&15.76\%&84.24\%\\
Baseline&10.94\%$_{\pm\textrm{3.38}}$&24.48\%$_{\pm\textrm{3.51}}$&35.16\%$_{\pm\textrm{2.92}}$&64.84\%$_{\pm\textrm{2.92}}$&19.53\%$_{\pm\textrm{3.55}}$&38.28\%$_{\pm\textrm{1.10}}$&54.43\%$_{\pm\textrm{1.84}}$&45.57\%$_{\pm\textrm{1.84}}$\\
Only Gaze&35.16\%$_{\pm\textrm{1.28}}$&70.05\%$_{\pm\textrm{1.33}}$&83.59\%$_{\pm\textrm{1.10}}$&16.41\%$_{\pm\textrm{1.10}}$&25.78\%$_{\pm\textrm{0.64}}$&47.66\%$_{\pm\textrm{3.19}}$&60.42\%$_{\pm\textrm{2.88}}$&39.58\%$_{\pm\textrm{2.88}}$\\
Only History and External&23.18\%$_{\pm\textrm{4.25}}$&37.50\%$_{\pm\textrm{4.47}}$&49.48\%$_{\pm\textrm{4.34}}$&50.52\%$_{\pm\textrm{4.34}}$&26.30\%$_{\pm\textrm{3.01}}$&47.14\%$_{\pm\textrm{5.12}}$&63.28\%$_{\pm\textrm{3.38}}$&36.72\%$_{\pm\textrm{3.38}}$\\
All&44.79\%$_{\pm\textrm{3.21}}$&67.45\%$_{\pm\textrm{4.10}}$&79.69\%$_{\pm\textrm{1.69}}$&20.31\%$_{\pm\textrm{1.69}}$&37.50\%$_{\pm\textrm{4.18}}$&59.90\%$_{\pm\textrm{4.83}}$&73.44\%$_{\pm\textrm{1.91}}$&26.56\%$_{\pm\textrm{1.91}}$\\
  \bottomrule
  \end{tabular}
  }
\end{table*}

We evaluate GestureGPT's performance by running the collected data through our system under four different context settings: 
1) the baseline with only the function list (\textit{Baseline}), 
2) with gaze information (\textit{Only Gaze}), 
3) with interaction history and external information (\textit{Only History and External}), and 
4) all contexts are available (\textit{All}).
For comparison, we also provide results from \textit{Random Guess}, which randomly selects one from all functions.
We repeat the tests three times under each setting to ensure robust conclusions.
The main results are presented in the left part of Table~\ref{tab:main_result}, and a corresponding illustration is provided in Figure~\ref{fig:home_task_results}.

Our first observation is that GestureGPT can effectively utilize the context information to determine the exact intention of the users.
The accuracy is effectively improved to an impressive extent when either gaze or history and external is incorporated.
When all contexts are combined together, the overall framework achieves the best Top-1 performance at 44.79\%.

When comparing the benefits of gaze versus history and external information, we empirically find that gaze significantly outperforms history and external information. 
As shown in Table~\ref{tab:main_result}, the \textit{Only Gaze} setting achieves 35.16\%/83.59\% at Top-1/Top-5 accuracy, outperforming the 23.18\%/49.48\% achieved by the \textit{Only History and External} setting by a considerable margin. 
We attribute this to the semantic similarity between candidate functions shared by different home devices and their broad spatial distribution across the entire room space. 
These two factors significantly increase the reasoning difficulty under the Smart Home scenario. 
Facing such challenging situations, GestureGPT can fully exploit the gaze information as well as its own commonsense knowledge to finally locate and zoom in on possible candidates through multi-step, collaborative spatial reasoning. 
As explained in Section~\ref{section:method_context_operations}, when the Gesture Inference Agent requests gaze information, the Context Management Agent not only outputs the gaze coordinates but also identifies the relevant device within the gaze path using external tool-augmented Python scripts.

\definecolor{topone}{HTML}{CC7E00}
\definecolor{topthree}{HTML}{FFA81A}
\definecolor{topfive}{HTML}{FFD899}
\definecolor{negative}{HTML}{91bfdb}

\begin{figure}[htbp]
  \vspace{-7pt}
  \centering
  \includegraphics[width=0.6\textwidth]{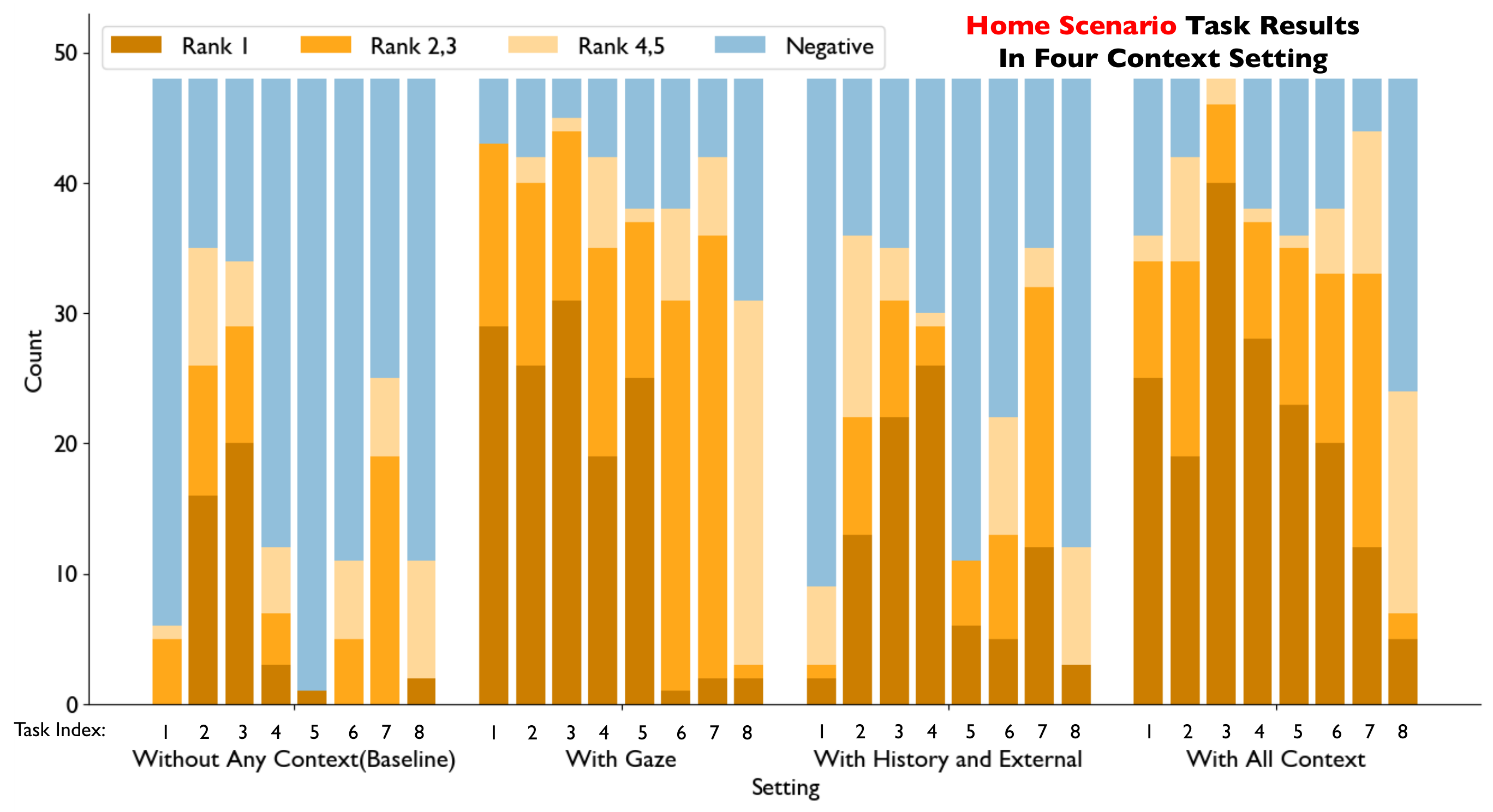}
  \caption{Illustration of results from Experiment 1: Home Scenario, encompassing all 4 settings (across the x-axis) and 8 tasks (indicated by hue). For each setting, the 8 tasks are arranged from left to right as follows: \textit{task 1 Unlock Carbinet}, \textit{task 2 Increase Light Brightness}, \textit{task 3 Next Recipes}, \textit{task 4 Open Oven}, \textit{task 5 Open Air cleaner}, \textit{task 6 Set Timer}, \textit{task 7 Switch Input}, and \textit{task 8 Phone Call}. Each specific bar is further divided into Top 1 / Top 3 / Top 5 accuracy (\textcolor{topone}{\textbf{Top 1}}, \textcolor{topthree}{\textbf{Top 3}}, \textcolor{topfive}{\textbf{Top 5}} in gradient brown) and Negative outcomes (\textcolor{negative}{in blue}), totaling 48 data points.}
  \label{fig:home_task_results}
  \vspace{-7pt}
\end{figure}

Moreover, while integrating all context achieves the highest Top-1 performance, it unexpectedly underperforms in the Top-5 metric compared to the Gaze Only setting (79.69\% vs 83.59\%). 
Through further case analysis, we hypothesize that an excess of context can sometimes disrupt the agent's analysis, leading to the following behaviors: 
1) Irrelevant context information can at times mislead the agent. For example, the mention of ``fingerprint unlocking'' (in task 1) within external information might lead the agent to infer fingerprint recognition as the unlocking method, rather than a gesture. 
2) An abundance of context causing the agent to overlook certain information. For example, in the ``Open the air cleaner'' task (in task 5) where the agent wrongly assumes the device is on, based on external information ``The air purifier's sensor detected heavy cooking fumes,'' while ignoring its actual OFF status.
These insights will be further discussed in Section~\ref{section:discussion_context_type} regarding optimal context selection.

Finally, we observe that Tasks 2, 3, and 7 exhibit relatively high baseline performance, all involving operations related to ``switching''. 
Finally, we observe that Tasks 2, 3, and 7 exhibit relatively high baseline performance, all involving operations related to ``switching''. 
Similarly, for Tasks 3 and 7, GestureGPT employed a comparable analytical approach to discern the correct answer. 
This demonstrates GestureGPT's strength in analyzing gestures alongside current device states, thereby accurately interpreting user intentions even in the baseline context.

In conclusion, our investigation into GestureGPT's performance across different contexts underscores the significant enhancements brought about by gaze data and the contributions of history and external information. 
Further refinement in how external information is presented and queried will unlock greater performance of our system. 

\subsection{Experiment 2: Online Video Streaming on PC}

This scenario is set when a user is watching online video on a PC monitor. 
Grounding gestures to the correct function is much more challenging in this scenario compared to the previous one. 
The video streaming interface contains a considerably broader range of functions, with numbers up to 66 in some tasks from the previous 18 functions. 
Moreover, many functions have similar semantic meanings, such as the ``vlog channel'' button and the ``anime channel'' button, making them difficult to distinguish solely through gestures. 
Furthermore, the interactive buttons and elements on the screen are much smaller than the smart appliances in the previous experiment, which reduces the performance of gaze-based function differentiation. 
The size of function elements ranges from \(0.27\) to \(20.71\) cm\(^2\) (\(\text{MEAN} = 2.73\), \(\text{SD} = 3.67\)).
By navigating through an interface rich in functions and semantic complexities, we intend to explore the boundary of our system's performance and understand whether it can differentiate user intentions with only minor differences, which is essential for real-world applications.

\subsubsection{Experiment Setting and Procedure}

\begin{itemize}
  \item \textbf{Experimental Platform and Gesture Data Collection} - Our experimental framework utilizes Python and Selenium\footnote{https://www.selenium.dev/} to interact with a video streaming platform, specifically targeting the website ``[China] From the Spring and Autumn Period to the Prosperous Tang Dynasty (Season 1, 12 Episodes)'' on Bilibili\footnote{\url{https://www.bilibili.com/video/BV1sh411j7A4/}}. The platform automates video control operations via Selenium. User gestures are captured using a 1080P resolution webcam.
  
  \item \textbf{Context Library Configuration} - The study incorporates a comprehensive context library comprising four distinct aspects same as in previous scenario:
  \begin{itemize}
    \item \textit{Interface Function List}: The function list is automatically extracted from the webpage and organized. Functions on the website included their position and raw HTML code are identified via JavaScript. The code for each function is then extracted and fed into GPT-4 to generate function names, aiding in the compilation of the interface function list. For tasks 4, 5, and 6, the function count is 17; for all other tasks, it is 66.
    \item \textit{Gaze Information}: The Tobii Eye Tracker 5 is used to collect gaze data from participants.
    \item \textit{Interaction History and External Context Information}: Interaction history was extracted from the task sequence, while the external contexts were predefined.
  \end{itemize}

    \item \textbf{Task Descriptions} - Eight tasks were designed to simulate common operations performed while watching videos (Table~\ref{tab:video_streaming_info}).

  \item \textbf{Participants} - We recruited 16 participants from four local schools, compensating them at a rate of \$12 per hour. Their ages ranged from \(18\) to \(35\) years (\(\text{MEAN} = 26.875, \text{SD} = 4.689\)), comprising \(10\) males and \(6\) females. Participants included 5 undergraduate students, 8 graduate students, and 3 research engineers. None of the participants had prior experience with AR/VR devices, which helped minimize bias due to varying familiarity with such technology.
  
  \item \textbf{Task Procedure} - Participants were briefed on the experiment's aims and the devices utilized upon their arrival. They were also asked to make any gesture deemed most intuitive with their right hand, similarly triggered by raising their hand above the chest. There was a warm-up phase for the participant to familiarize with gesture controls. Following this, the participant sequentially completes the eight tasks. Specific gestures triggered predefined responses on the website, simulating real-time interaction for a more realistic experience. 
\end{itemize}

As a result, a total of \(16 \text{ participants} \times 8 \text{ tasks} = 128 \text{ data points}\) were collected. Detailed information about the experiment simulation interface, the list of devices and their functions, and the task list is provided in Appendix~\ref{appendix:exp_setting_video}.

\subsubsection{Results Analysis}

\begin{figure}[htbp]
  \centering
  \includegraphics[width=0.6\textwidth]{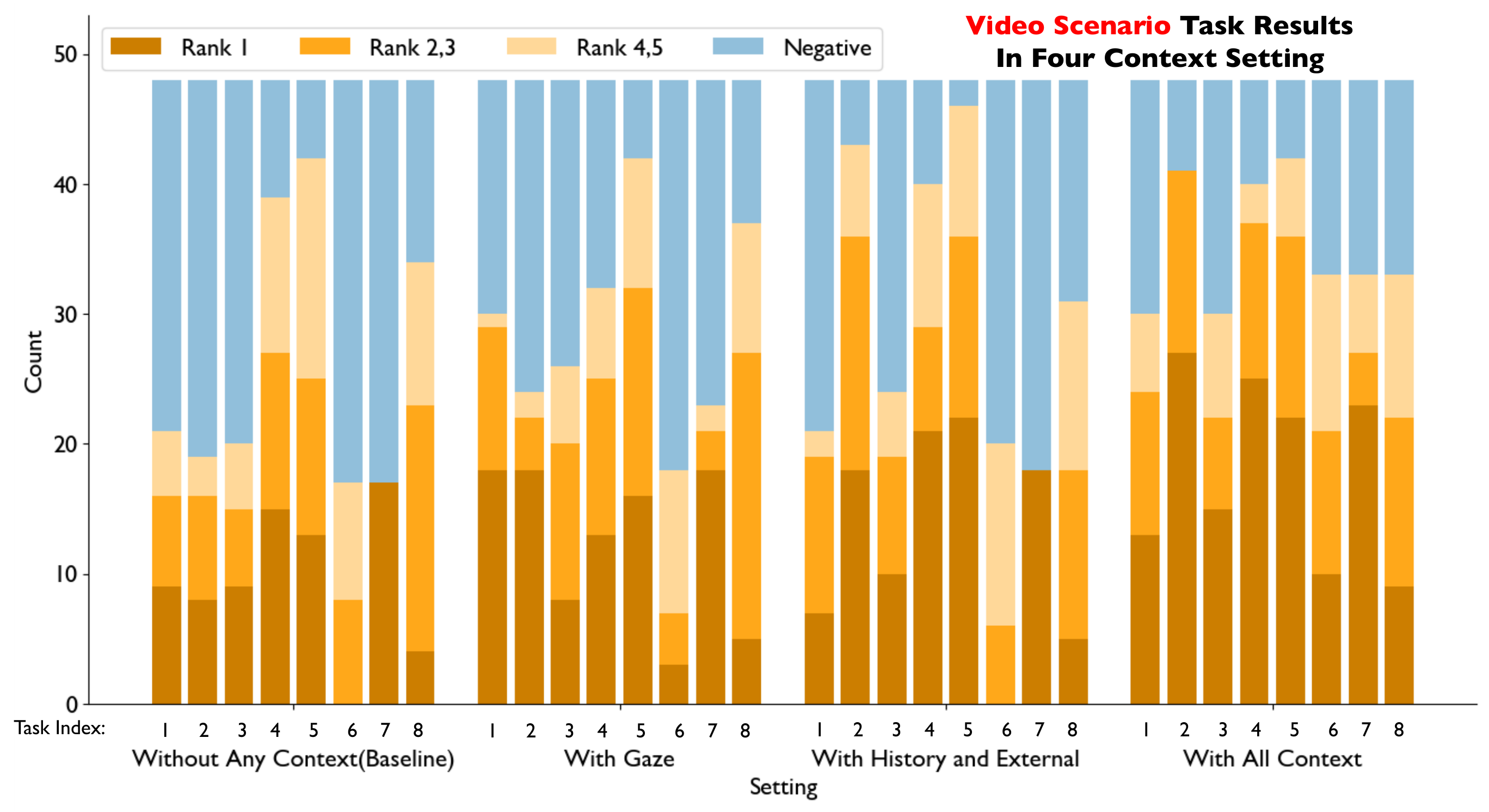}
  \caption{Illustration of results from Experiment 2: Video Scenario, encompassing all 4 settings (across the x-axis) and 8 tasks (indicated by hue). For each setting, the 8 tasks are arranged from left to right as follows: \textit{task 1 Adjust Volumn}, \textit{task 2 Video Progress Control}, \textit{task 3 Enter Fullscreen}, \textit{task 4 Pause Video}, \textit{task 5 Play Video}, \textit{task 6 Exit Fullscreen}, \textit{task 7 Like the Video}, and \textit{task 8 Next Video}. Each specific bar is further divided into Top 1 / Top 3 / Top 5 accuracy (\textcolor{topone}{\textbf{Top 1}}, \textcolor{topthree}{\textbf{Top 3}}, \textcolor{topfive}{\textbf{Top 5}} in gradient brown) and Negative outcomes (\textcolor{negative}{in blue}), totaling 48 data points.}
  \label{fig:video_task_results}
\end{figure}

In this scenario, we evaluate GestureGPT under the same four context settings.
The results are presented in the right part of Table~\ref{tab:main_result} as well as Figure~\ref{fig:video_task_results}.

The main conclusion that GestureGPT can effectively incorporate various context information to reason and predict user intentions remains consistent. 
Specifically, Top-1 accuracy improved from 19.53\% to 37.50\% when all types of context are available, significantly outperforming each single context setting: +11.20\% compared to only the history and external setting, and +11.72\% compared to the only gaze setting. 
Both contexts contribute to the final performance, and it greatly drops if either is excluded. 
Compared with Experiment 1, where gaze information brings relatively more utility, these results further demonstrate that it requires joint reasoning over all available contexts to achieve the best performance under such a more complicated scenario.

An intriguing observation from our results was that the baseline performance was superior to that observed in the home scenario, thereby necessitating task-specific exploration. 
Tasks such as pausing, playing videos, and selecting the next video (Tasks 4, 5 and 8), which require less context, exhibited strong performance across all tested context settings. 
This phenomenon suggests that the agent's inherent knowledge of video streaming interfaces makes it likely to infer actions commonly seen in such a video streaming interface. 
To some extent, the agent's success in identifying these operations can be attributed more to an educated guess rather than a calculated match of gesture and context, indicating a form of intuition derived from the model's extensive training data.
This intuition helps our system's performance in certain tasks, but also acts as bias in other tasks. 
This phenomenon is discussed further in Section~\ref{sec:llm_bias}. 

Another special case is Task 7, which is semantic specific task, highlights the essential role of accurate gesture description in achieving precise gesture grounding. 
Out of 16 participants, 12 performed the ``thumbs-up'' gesture for the task. 
In such cases, if the agent accurately recognizes the gesture, the outcome is correct, notably reflecting in high Top-1 accuracy results.
The failures predominantly occurred due to incorrect gesture segmentation. 
For example, because users formed a fist after lifting their hand, leading the agent to mistakenly focus on the action of making a fist, which leads to map the ``Full Screen'' or ``volume'' function. 
We conducted a pilot validation exercise on the gesture segmentation rule used in our system with a human labeler splitting gesture frames from the whole video on data from eight participants as the ground truth in gesture segmentation. Our results, compared to the ground truth, revealed a high recall rate of 95.28\%, yet a precision of only 43.91\%, reflecting the inclusion of non-gesture-related frames. Future enhancements could involve advanced object detection techniques , differentiating interactive vs non-interactive gestures, or a specially designed beginning gesture to improve segmentation precision.

In cases where gesture descriptions were poorly articulated, incorporating additional context proved beneficial. 
For example, considering interaction history context, such as users exiting full-screen mode, allowed the agent to infer that the user might want to perform operations unrelated to video control, like ``liking'' a video. 
This various context adaptation significantly improved this task performance in Top-5 accuracy metrics in this task from 35.42\% to 68.75\%.

Despite the challenges we designed in the video streaming scenario, GestureGPT still demonstrates commendable performance, largely attributed to the LLM's robust common sense and contextual interpretation capabilities.

\subsection{Result Analysis: Gesture Description Quality Assessment}

The quality of the Gesture Description Agent was evaluated through an expert questionnaire. 
For this evaluation, we randomly sampled descriptions generated from three repetitions for 256 gestures across two scenarios, resulting in a compilation of 256 gestures and their descriptions.
The same questionnaire was then distributed to two gesture experts, who rated each gesture description on a 5-point Likert scale (1 being ``Almost no description is correct or relevant to the gesture'' and 5 being ``Almost all of the description is correct or relevant to the gesture''). Both expert have published gesture-related research articles on premier conferences. 
One expert scored 3.28 (std = 1.41), and the other scored 3.74 (std = 0.73). Their Kendall's W score was 0.853, indicating a high level of agreement between the two experts.
The positive ratings show that our description agent can capture key information of the gesture. 

\subsection{Pilot Study: Human Performance in Gesture Understanding}

\begin{figure}[htbp]
  \vspace{-10pt}
  \centering
  \includegraphics[width=0.5\textwidth]{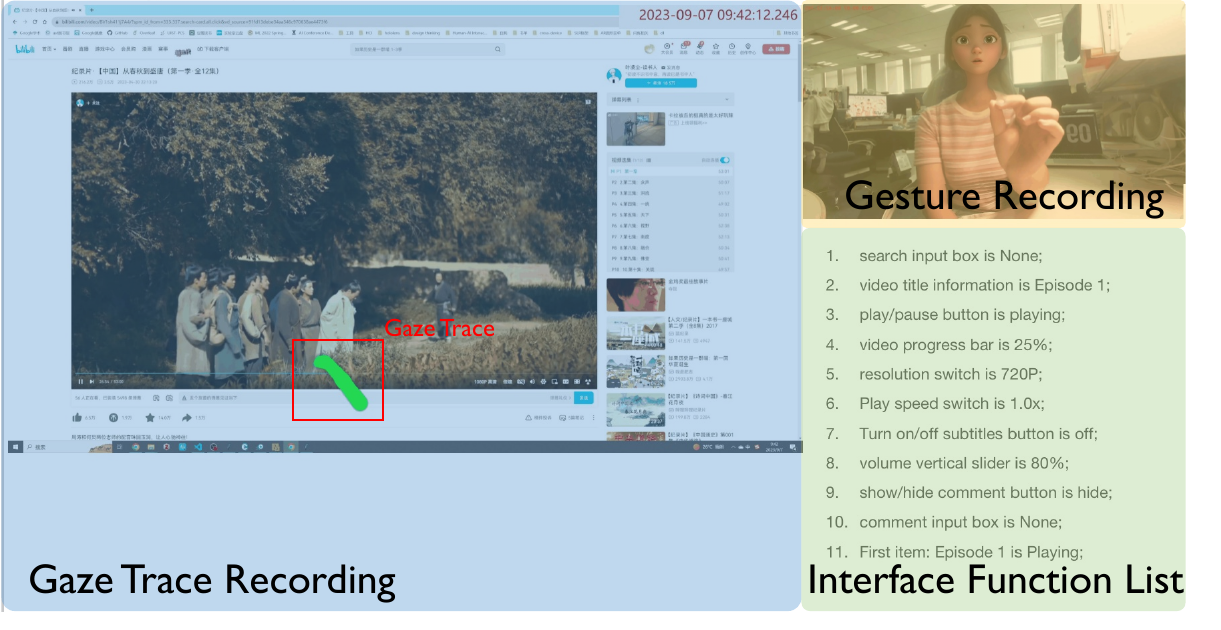}
  \caption{Pilot study interface showcasing. On the left, a video stream with a highlighted gaze trace indicating user focus during the task. On the right, a gesture recording from a video streaming scenario alongside a detailed interface function list. This setup was used to assess participant ability to map gestures to corresponding interface functions, facilitated by context such as gaze trajectories and historical information.}
  \label{fig:pilot_interface}
  \vspace{-10pt}
\end{figure}

To gauge human performance on tasks similar to those managed by GestureGPT, we conducted a pilot study with four participants who were unfamiliar with the system and its experimental setups.
We randomly selected data from four users involved in Smart Home and Video Streaming scenarios.
During the pilot, participants were provided with video recordings that included the user's gesture poses and movements, along with a list of interface functions available at that time. They were tasked with mapping each gesture to its corresponding function.
The pilot study interface as shown in~\ref{fig:pilot_interface}.

Initial attempts by participants were guided by \textit{gaze context}, as demonstrated through gaze trajectory recordings, and \textit{historical context}.
Subsequent attempts were further supported by providing \textit{external context}, the pilot study interface shown in~\ref{fig:pilot_interface}.
Subsequent attempts were further supported by providing \textit{external context}. 
Results revealed significant challenges in gesture understanding even for humans: initial Top-1 and Top-5 accuracies were 43.8\% and 81.3\% for the smart home, and 18.8\% and 50\% for video streaming, respectively. 
With added external context, accuracies increased notably to 58.3\% (Top-1) and 91.7\% (Top-5) in the smart home, and to 83.3\% (Top-1) and 100\% (Top-5) in video streaming. 
This preliminary result underscores the inherent complexity of interpreting gestures, even for humans, in contextually rich settings.
So compared with gesture recognition, gesture understanding is much more difficult and challenging.

\section{Discussion}
\label{sec:discussion}

\subsection{Integration and Utilization of Top-N Predictions}
\label{sec:selection_method}
Due to the inherent complexity of gesture understanding—far surpassing that of mere gesture recognition—we approach this challenge as a recommendation task, which is why we show the Top-3 and Top-5 accuracies.
GestureGPT currently outputs a Top-5 candidate function list, which can be readily augmented with list selection interaction optimizations, as studied in many existing works.
Specifically, \citet{noauthor_dwell_nodate} developed a dwell selection system that employs machine learning to predict user intent solely based on eye movement data. 
Furthermore, statistical design of dynamic menus could also expedite user selection particularly in AR/VR contexts. 
G-Menu~\cite{noauthor_g-menu_nodate} utilizes gestures and keywords to design a dynamic menu, aiding in swift function selection. 
These techniques and designs can be readily integrated into GestureGPT to achieve efficient user selection of the target function. 
The rapid advancement of LLMs and their increasing capability as contextualized agents, LLM-driven frameworks like GestureGPT will provide improved performance so that its Top-1 accuracy reaches a directly applicable level, ultimately ease the burden of secondary selection of users.

\subsection{Language Input (LLM) vs. Visual Input (LMM) for Agents}
\label{dis:lmm}
GestureGPT is currently built with a Large \textit{Language} Model for its superior understanding and reasoning capability. For the Gesture Description Agent, an alternative implementation could employ a Large Multimodal Model (LMM), which seems more intuitive. So we tested replacing the rule-based Gesture Description Agent with GPT-4o's vision in a video streaming scenario. With \textit{All Context} considered, the Top-1 and Top-5 accuracy results were 41\% and 72\%. These results demonstrate performance comparable to our rule-based module which is 37.5\% and 73\%, thereby validating the effectiveness of the rule-based approach.

On the other hand, LMMs require users to upload gesture recordings, which raises further privacy concerns. By contrast, our solution can process the gesture input with end-affordable devices and only send anonymized skeleton signals to the data center, where LLMs can perform dense computing. Nevertheless, it is anticipated that with further development of LMMs, GestureGPT could integrate both the visual recognition capabilities of LMMs and the prominent commonsense understanding, contextual reasoning capabilities of LLMs, to ultimately better serve its purpose.

\subsection{Single Agent vs. Multi Agent}
\label{sec:dis_singleagent}

Assuming privacy concerns regarding LMMs are not considered, employing gesture visuals with contextual information in a LMM-based single-agent architecture could significantly reduce the system delay.
Unlike multi-agent systems, which require multiple reasoning cycles to interpret gestures, a single-agent approach can deliver immediate results within one iteration.
We conducted a preliminary test with data from one user in the smart home scenario, and the LMM-based single-agent approach achieved a Top-1 accuracy of 25.0\% and a Top-5 accuracy of 75\%.
Due to privacy concerns, comprehensive testing was limited because other users did not authorize the publication of data.

Although the end-to-end LMM approach exhibits shorter delays, it presents several disadvantages: it is cumbersome to adjust and train, lacks task transferability, and falls short in terms of scalability and generality.
Conversely, the multi-agent architecture offers significant advantages, such as enabling component reuse, and facilitating modular training and optimization.
Each architecture has its own applicable scenarios, and together, they can complement each other to achieve superior results.

However, the multi-agent architecture of GestureGPT inherently offers greater flexibility and is designed for broad applications.
For example, instead of cameras, the description agent can work with sensors like IMUs, electromyography, and LiDAR, provided they can reconstruct the hand pose to some extent.
The framework handles complex interaction contexts to mimic human gesture understanding, such as interpreting gestures linked to ongoing events.
Integrating context into the inference agent would greatly increase the workload and limit flexibility.
The independently implemented modules allow for easier optimization.
Smaller models within the architecture are plug-and-play, thereby avoiding constant end-to-end training.

Furthermore, we contend that the multi-agent architecture of GestureGPT is ideally suited for complex interaction scenarios where single-step reasoning is insufficient.
For example, an agent might need to interact with users or external databases before making decisions.
Our architecture supports distributed processing, enabling different agents to handle specific tasks asynchronously, thereby having the potential to further enhance efficiency in managing complex interactions.
For a detailed discussion on how this architecture addresses system delays, see Section~\ref{sec:limitation_delay}. 

\subsection{Context Selection in Complex Systems}
\label{section:discussion_context_type}

Our experiments revealed that although adding more contexts is supposed to be informative and useful, for certain circumstances, it might also bring overwhelming irrelevant information and accordingly distract the reasoning process of LMM agents (See Top-5 accuracy of \textit{All} setting vs \textit{Only Gaze} setting in Smart Home scenario).
In Smart Home scenario, most informative contexts are concentrated on gesture itself and gaze, while history and external provide relatively less contribution to the final performance.

On one hand, the inclusion of more heterogeneous contexts increases the complexity and difficulty of context organization and management, which then requires more competent agents to process and reason over them.
On the other hand, contexts with less information entropy inevitably brings irrelevant implications, and may result in misleading of agents for specific cases.
This underscores the importance of making informed trade-offs between context incorporation and agent capability to optimize the system performance.

\subsection{LLM Common Sense Bias}
\label{sec:llm_bias}
This phenomenon pertains to the cognitive biases of LLM,
which has been recognized as a common issue that influences LLM outputs~\cite{samsi_words_2023}.

Such a bias was also observed in our system, notably within the video streaming scenario. 
When the agent learned that this is a video streaming scenario, there is a bias towards predicting video control functions that are commonly used on such an interface, such as play, pause, and volume control. 
On one hand, even in the absence of context, the agent can make accurate guesses if the intended function is one of such functions. 
On the other hand, it leads to the consistent inclusion of these functions within the candidate options, detrimentally impacting the top-1 accuracy.

One way to address this bias could involve employing Modular Debiasing Networks~\cite{gallegos_bias_2024} to mitigate bias or utilizing sophisticated prompt engineering to diminish its effects, which we plan to investigate in the future. 

\subsection{System Scalability}

By substituting the Gesture Description Agent with specially designed counterparts, our system can adapt to more input modalities and more generalized form of gestures.

In our implementation, gesture feature extraction is solely based on a RGB camera and MediaPipe. 
Yet this approach is susceptible to lighting conditions and finger occlusion issues. 
Wearable devices provide an alternative robust solution for hand reconstruction~\cite{zhang_ultraglove_2023, hu_fingertrak_2020, xu_enabling_2022}, which even works with subtle gestures like thumb-tip gestures~\cite{gong_pyro_2017} and wrist gestures~\cite{gong_wristwhirl_2016, gong_wristext_2018}. 
Hand landmarks reconstructed from wearable sensors can then be used to generate gesture description. 

Our system can also be extended to general gestures, beyond the scope of merely hand gestures. For example, touch-screen gestures can be extracted as traces and pressure intensity, which differ significantly from the form of hand gestures. But a specially designed Gesture Description Agent can extract the features of such gestures (either using rule-based methods or leveraging large language or vision models) and describe it to Gesture Inference Agent, thus easily integrated into our framework.

\section{Limitation and Future Work}

\subsection{Handling Left-Handed Gestures, Multiple Hands, and Non-Interactive Gestures in Frame}
\label{sec:lim_hand}

GestureGPT currently identifies meaningful gestures when a right hand is raised above chest level. 
However, this implementation assumes that only one hand, typically the right hand, is visible.
This assumption is often unrealistic, as interactions in real environments may involve left-handed users or require two-handed gestures, thereby limiting the system's inclusivity.
Furthermore, this simplification can lead to false activations, as unintended gestures or non-interactive movements may inadvertently trigger the system.
This issue is exacerbated in scenarios where various non-interactive hand movements and multiple hands are present within the camera frame, potentially confusing the model and degrading system performance.
Evaluating the system's robustness when both hands are visible or when left-handed gestures are involved is essential for fully understanding its practical capabilities.
Future iterations of our system will focus on distinguishing between intentional gestures and extraneous hand movements, even in the presence of multiple hands, and on improving left-hand gesture recognition to better accommodate left-handed users.

Several existing studies have addressed these challenges. Research~\cite{Just2006} has explored techniques for distinguishing between the static and dynamic aspects of hand movements, along with the challenges of recognizing one- and two-handed gestures. 
Similarly, vision-based interaction approaches for augmented reality~\cite{szalavari1997personal} emphasize the importance of simplifying hand models and applying user intention recognition methods to achieve robust performance in complex scenarios.
Additionally, frameworks that incorporate body pose, gaze direction, and multimodal feedback mechanisms have shown promise in predicting interaction intent~\cite{schwarz_combining_2014, freeman_that_2016}, while other studies have integrated EEG and gaze patterns to differentiate between meaningful and irrelevant gestures during tasks~\cite{ge_improving_2023, koochaki_predicting_2018}.

The architecture of GestureGPT is designed to be adaptable and scalable, enabling the integration of these advancements to further enhance its capabilities. Incorporating a left-hand recognition module within the Gesture Description Agent would allow the system to support both left-hand and two-handed gestures. Additionally, expanding the contextual input processed by the Context Management Agent could improve the system’s ability to distinguish non-interactive gestures, thus reducing the likelihood of false activations.
In the future, we plan to explore these challenges further, including validating the accuracy of left-hand gesture recognition, providing comprehensive support for two-handed gestures, and refining the system’s activation mechanisms to enhance its practicality and robustness.

\subsection{System Delay and Applicability}
\label{sec:limitation_delay}

The overall time cost of the framework is primarily driven by LLM-related processes, accounting for up to 90\% of the total time expenditure.
These costs include LLM API request costs. For each gesture understanding task, GestureGPT typically requires 1 request for the gesture description and 6 rounds of conversation between the Gesture Inference Agent and Context Management Agent, highlighting the significant impact of response delay on the real-time performance of our system.
In its current conceptual form, without a focus on practicality, our framework averages 227 seconds and 38,785 tokens per task.

Although the conceptual framework of GestureGPT demonstrates promising potential for HCI applications, the current implementation faces significant challenges related to system delays, which affect its real-world applicability.
Addressing these delays is essential for transitioning GestureGPT from a theoretical model to a practical tool capable of functioning in dynamic, real-time environments.
Possible directions for improvements include:
\begin{itemize}
    \item \textbf{KV-cache}: API calls require concatenating long system prompts in each round, exponentially increasing time costs. Our current implementation appends the entire context library (11,000 tokens) to each query. Implementing a KV-cache would reduce the need to reprocess identical contexts, potentially decreasing interaction time to approximately 20 seconds per task. This enhancement is expected to streamline operations by preserving computed attention values for recurring queries.
    \item \textbf{High-Performance Hardware Utilization}: Integrating advanced computational resources, such as Groq's hardware, could further reduce processing times. Groq's fast model inference services offer an output speed of over 700 tokens per second.
\end{itemize}
By combining Groq and KV-cache, the first context round would take 2-3 seconds, with subsequent rounds taking a total of about 1-2 seconds (averaging 6 dialogue rounds, each with 200 tokens input and 100 tokens output). This would reduce the theoretical optimal latency to under 5 seconds. Although we do not currently have access to KV-cache-enabled high-performance LLMs or the computing acceleration hardware provided by Groq, we anticipate that GestureGPT will attract interest from other research domains, fostering collaborative efforts towards developing a practical and accessible free-form gesture interface.

On the other hand, since each agent of our system has specialized and clear goals, we believe a properly fine-tuned LLM could mirror GPT-4's effectiveness with fewer resources as demonstrated in NLP tasks~\cite{zhong_can_2023}.
Besides, advances in model distillation and acceleration hardware suggest that lightweight LLMs could be deployed directly on devices with fast inference~\cite{zhao_lingualinked_2023,zhao_lingualinked_2023,ma_poster_2023} for real-time performance in the future. 
Under which circumstances, the API request cost can also be removed.

While the current system delays may limit immediate real-world applications, the outlined advancements and potential collaborations offer a promising path toward minimizing these delays. 
Future iterations of our system, supported by ongoing research and technological improvements, could effectively reduce latency to levels suitable for dynamic, real-time user interactions. 
Such developments would not only address the practical limitations discussed but also pave the way for broader adoption and impact across various domains, ultimately fulfilling the potential of real-time, intuitive gesture-based interfaces.

\section{Conclusion}

In this work, we introduce GestureGPT, a human-mimicking framework for understanding free-form hand gestures, which capitalizes on the capabilities of LLMs. Unlike traditional gesture recognition systems that require predefined gestures, GestureGPT automatically maps spontaneously performed, free-form gestures to their targeted interface functions without any user training. This is achieved through a collaborative framework involving three specialized agents: a Gesture Description Agent, a Context Management Agent, and a Gesture Inference Agent. These agents work together to interpret gestures, manage context, and apply common sense reasoning to accurately discern user intents.

Our offline evaluations in two realistic scenarios demonstrated promising results, achieving the highest zero-shot Top-5 gesture grounding accuracy of 83.59\% for smart home control and 73.44\% for video streaming. These outcomes highlight the significant potential of GestureGPT as a conceptual model for future gesture-based interaction systems. They underscore the viability of our approach in reducing the cognitive load on users by eliminating the necessity to learn, memorize, or manually link gestures to functions.

As a conceptual framework, GestureGPT sets the groundwork for future advancements in natural interaction interfaces. While not yet a practical system ready for everyday use, our framework serves as a foundational step towards realizing more adaptive and intuitive ways to interact with technology. Future research could focus on refining this paradigm, enhancing its responsiveness, and expanding its applicability to include a broader array of natural user interfaces supporting different modalities like facial expressions, body movements, and postures, unleashing the full potential of the framework proposed by GestureGPT.

\section*{Acknowledgments}

The authors would like to thank Benfeng Xu, whose insightful comments and suggestions were invaluable in writing this paper. 
This work was supported by the Hunan Provincial Natural Science Foundation of China (No. 2023JJ70009), the HNXJ Philanthropy Foundation (KY24017), the Science and Technology Innovation Program of Hunan Province (No. 2022RC4006), and the Young Scientists Fund of the National Natural Science Foundation of China under Grant No. 62102401.

\appendix

\section{Gesture Description Rules}
\label{apx:rule}

\subsection{Rules Calculation Method}
\label{apx:rule_calculation_description}
\textbf{Flexion} of a finger is calculated as the total bending angle of each joint. For thumb it is the bending angle of the ip joint, and for other fingers, it is the bending angle of the pip and dip joint. Then, two parameters $straight\_threshold$ and $bent\_threshold$ are set to determine if the finger is straight, bent, or ``unsure'' if the result falls between them. 
Since thumb has a different joint structure compared with other fingers, a new pair of thresholds are specially set for thumb.

\textbf{Proximity} of two fingers is calculated as the average minimal distance from each finger's joint to the other finger. 
Two thresholds \ie $together\_threshold$ and $separated\_threshold$ are set to determine if the two fingers are pressed together, separated, or ``unsure'' if the result falls between them.

\textbf{Contact} of thumb and another finger is computed as the distance between their fingertips. Then, two thresholds, \ie $contact\_threshold$ and $not\_contact\_threshold$ are set to determine if the two fingers' fingertips are in contact or not, or ``unsure'' if the result falls between them.

\textbf{Point direction} of thumb is computed as the direction from thumb's mcp joint to tip joint. Then, it is compared with two reference vectors representing upward and downward. If a reference vector has the minimal angle with the palm orientation vector and the angle is below $angle\_threshold$, the reference vector would be the thumb's pointing direction. Note that it is only applicable when thumb is straight. If thumb is bent, the result is set to ``unsure''. This is especially useful when discriminating between gestures like ``thumb up'' and ``thumb down''.

\textbf{Palm orientation} is computed as the direction to which the palm is facing. It is computed by the cross product of two vectors within the plane of the palm (Figure~\ref{fig:rule_calculation}). Then, the direction is compared with six reference vectors representing upward, downward, left, right, inward and outward. If a reference vector has the minimal angle with the palm orientation vector and the angle is below $angle\_threshold$, the reference vector would be the palm orientation. Otherwise the orientation of palm is set as ``unsure''.

\textbf{Hand position} is computed as the geometrical center of a hand by taking average of all 21 landmarks' coordinates. No parameter or threshold is applied here.

\subsection{Training for Threshold and Evaluation for Gesture Description Rules}
\label{apx:rule_training_and_evaluation}

We use HaGRID~\cite{kapitanov_hagrid_2024} dataset to tune and evaluate our rules. HaGRID contains images of 18 kinds of gestures (call, dislike, fist, four, like, mute, ok, one, palm, peace, peace inv., rock, stop, stop inv., three, three 2,  two up, two up inv.). 

\subsubsection{Training Process}

A subsample split of HaGRID has 100 images per gesture, and it is used to tune the rule parameters. 

To tune the rule parameters, we first manually annotate the ground truth label for each rule on each gesture class. 
For each gesture, we assume that most people perform it in the same way, and thus the ground truth label is obtained by common knowledge. (For example, for ``thumb up'' gesture, we label thumb as straight, index to pinky fingers as bent, index and middle finger are pressed together, \etc)
 
However, there exist ambiguous cases, in which more than one labels are acceptable. (For example, in ``peace'' gesture, it is reasonable for thumb to be either straight or bent.) In this case, the label is more than one, and consequently cases like this are not used for parameter tuning. 

Most rules are tuned using the whole subsample split. One exception is \textbf{Finger Flexion} on thumb, because in most gesture classes, the ground truth of thumb is either [``straight''] or [``straight'', ``bent''] (thus not used for training), and only one class is [``bent'']. To address the issue of severe imbalance for thumb flexion, we specially select the training set, which is composed of gesture ``like'' (thumb is straight) and gesture ``ok'' (thumb is bent), and each image in this set is manually checked to ensure the quality of the training data.

As is shown above, for those cases used to tune the rules, the ground truth label is one of the two possible states (\eg ``straight'' or ``bent''). But during prediction, we use Three-Way Decision Making method~\cite{hutchison_outline_2012}. When the rule cannot clearly determine the state, it will predict it as ``unsure'', to avoid wrong conclusions that may easily mislead LLMs. Consequently, different from the classical binary classification, some modifications have been made to the correctness assessment and tuning criteria to accommodate our specific setting.

\textbf{Correctness Assessment}. For each (prediction, label) pair, there are three possible circumstances: 
\begin{itemize}
  \item Unsure: Prediction $=$ ``unsure''.
  \item Error: Prediction $\neq$ ``unsure'', and prediction $\neq$ label.
  \item Correct: Prediction $\neq$ ``unsure'', and prediction = label.
\end{itemize}

\textbf{Tuning criteria}. To find the optimal parameters, we define the loss in different cases as follows:
\begin{itemize}
    \item Unsure: Loss = 0.2
    \item Error: Loss = 1
    \item Correct: Loss = 0
\end{itemize}
In ``unsure'' cases, the loss is between 0 and 1. This is because if we set the loss to be 0 (same as ``correct'' case), then the rules tend to predict every case to ``unsure'; if we set the loss to be 1 (same as ``error'' case), then the rules tend to not predict any ``unsure'', which may lead to misleading predictions.

A grid search is conducted to find out the optimal parameter with minimal average loss. The optimal parameters for each rule is shown in Table~\ref{tab:Parameters and Performance}.

\subsubsection{Test Performance}

We test the generalization ability of our rules on two dataset: HaGRID test set (third-person view) and EgoGesture dataset (first-person view)~\cite{zhang_egogesture_2018}.

The performance on HaGRID test set (38576 images) is shown in Table~\ref{tab:Parameters and Performance}.
For most rules, the average error rate among all gestures is below 4.2\% and the overall accuracy is above 88.2\%. 
One exception is flexion rule for thumb, in which the output ``unsure'' takes about 45\% of all cases. 
This may be attributed to the unique shape of thumb: the topmost segment of the thumb is curved by nature, so when the thumb is extended, the landmark seems to be slightly bent, which may affect the train and test process. But by a three-way decision,
we use ``unsure'' to avoid the misleading information that may potentially confuse the Gesture Description Agent. 

When checking the algorithm's performance for each gesture, it is shown that the error rates for most gestures are below 5\% and none of them are above 16\%. 
The error could be attributed to landmark mistakes made by MediaPipe, the shortcomings of our algorithm, and a small number of people just perform the gesture differently from most people (\ie different from the ground truth we established before).
The results show good generalizability since the parameters are only tuned on a small number of samples. 

\begin{table}[htbp]
  \caption{Rule Parameters, and Their Performance on HaGRID Test Set}
  \label{tab:Parameters and Performance}
  \begin{tabular}{|c|c|c|c|c|}
    \hline
    \multirow{2}{*}{\textbf{Rule}} & \multirow{2}{*}{\textbf{Parameters}} & \multicolumn{3}{c|}{\textbf{Performance on HaGRID Test Set}}\\
    \cline{3-5} 
    &  & \textbf{error} & \textbf{unsure} & \textbf{correct} \\
    \hline
    Flexion - thumb & (16, 38) & 0.036 & 0.457 & 0.507 \\
    \hline
    Flexion - other fingers & (57, 74) & 0.019 & 0.049 & 0.932 \\
    \hline
    Proximity & (0.024, 0.029) & 0.031 & 0.067 & 0.902 \\
    \hline
    Contact & (0.046, 0.055) & 0.020 & 0.024 & 0.956 \\
    \hline
    Pointing Direction - thumb & 40 & 0.047 & 0.239 & 0.714 \\
    \hline
    Palm Orientation & 41 & 0.042 & 0.075 & 0.882 \\
    \hline
    Overall & - & $0.023\pm0.018$ & $0.062\pm0.039$ & $0.916\pm0.053$  \\ 
    \hline
  \end{tabular}
\end{table}

To evaluate if the parameters trained on third-person view dataset can adapt to first-person view images, we tested them on a first-person view gesture dataset EgoGesture. We choose 20 gesture classes (fist, measure, zero, one, two, three, four, five, six, seven, eight, nine, ok, three2, C, thumb down, thumb right, thumb left, thumb backward, thumb forward) and label the ground truth for each class. Each gesture has around 250 testing samples. The results are shown in Table~\ref{tab:Performance on EgoGesture Dataset}. (Thumb pointing direction is not evaluated on this dataset because there is no adequate gesture for testing. Only the ``thumb down'' gesture has a clearly downward pointing direction, yet the images are not pointing strictly downward.)

On this first-person view dataset, the error rate of the rules only increase from 2.3\% to 6.3\% (though the correct rate decrease by around 19\% because more ``unsure'' are predicted), showing that our rules works across different views. 

\begin{table}[htbp]
  \caption{Performance of Rules on EgoGesture Test Set}
  \label{tab:Performance on EgoGesture Dataset}
  \begin{tabular}{|c|c|c|c|}
    \hline
    \multirow{2}{*}{\textbf{Rule}} & \multicolumn{3}{c|}{\textbf{Performance on EgoGesture}}\\
    \cline{2-4} 
    & \textbf{error} & \textbf{unsure} & \textbf{correct} \\
    \hline
    Flexion - thumb & 0.046 & 0.532 & 0.422 \\
    \hline
    Flexion - other fingers & 0.060 & 0.218 & 0.722 \\
    \hline
    Proximity & 0.098 & 0.194 & 0.708 \\
    \hline
    Contact & 0.044 & 0.159 & 0.797 \\
    \hline
    Pointing Direction - thumb & - & - & - \\
    \hline
    Palm Orientation & 0.108 & 0.250 & 0.642 \\
    \hline
    Overall & $0.063\pm0.043$ & $0.216\pm0.096$ & $0.720\pm0.123$ \\
    \hline
  \end{tabular}
\end{table}

More detailed error analysis of our rules' performance on EgoGesture Dataset can be found in Table~\ref{tab:Error Analysis on EgoGesture Dataset}. Only those with error rate above 15\% are shown.

\begin{table*}[t!]
  \centering
  \caption{Analysis of Error Cases on EgoGesture Dataset}
  \label{tab:Error Analysis on EgoGesture Dataset}
  \resizebox{\columnwidth}{!}{
  \begin{tabular}{|m{1.7cm}|m{1cm}|m{2cm}|m{1.5cm}|m{7cm}|}
    \hline
    \textbf{Rule} & \textbf{Gesture} & \textbf{Finger} & \textbf{Error Rate} & \textbf{Observed Reasons} \\
    \hline
    \multirow{6}{*}{\parbox{3cm}{Flexion -\\ other fingers}} & seven & ring & 0.171 & MediaPipe's mistake for occluded fingers. \\
    \cline{2-5}
    & \multirow{2}{*}{C} & ring & 0.595 & The finger in this gesture is slighly bent by nature, hard to predict precisely. \\
    \cline{3-5}
    &   & pinky & 0.360 & Same as above. \\
    \cline{2-5}
    & thumb down & \multirow{2}{*}{index} & 0.177 & MediaPipe's mistake for occluded fingers. \\
    \cline{3-5}
    & & ring & 0.326 & Same as above. \\
    \cline{2-5}
    & three-2 & ring & 0.281 & Same as above. \\
    \hline    
    \multirow{11}{*}{Proximity} & C & middle-ring & 0.255 & The rule does not generalize very well. \\
    \cline{2-5}
    & three & middle-ring & 0.154 & Same as above. \\
    \cline{2-5}
    & \multirow{3}{*}{four} & index-middle & 0.260 & Landmarks mistake; some people perform it differently; the fingers are slightly separated by nature, hard to predict precisely, but it doesn't influence the recognition of the gesture very much.\\
    \cline{3-5}
    &  & middle-ring & 0.536 & Same as above.\\
    \cline{3-5}
    &  & ring-pinky & 0.353 & Same as above.\\
    \cline{2-5}
    & \multirow{3}{*}{five} & index-middle & 0.289 & Same as above.\\
    \cline{3-5}
    &  & middle-ring & 0.767 & Same as above.\\
    \cline{3-5}
    &  & ring-pinky & 0.488 & Same as above.\\
    \cline{2-5}
    & \multirow{2}{*}{ok} & middle-ring & 0.404 & Same as above. \\
    \cline{3-5}
    &  & ring-pinky & 0.578 & Same as above. \\
    \cline{2-5}
    & nine & index-middle & 0.244 & Landmarks mistake. \\
    \hline
    \multirow{7}{*}{Contact} & \multirow{2}{*}{seven} & thumb-ring& 0.202 & Some people perform it differently; MediaPipe's mistake for occluded fingers. \\
    \cline{3-5}
    & & thumb-pinky & 0.151 & Same as above. \\
    \cline{2-5}
    & measure & thumb-index & 0.198 & Landmark mistake; in some gestures thumb and index finger are close so it is hard to discriminate. \\
    \cline{2-5}    
    & \multirow{2}{*}{nine} & thumb-index & 0.191 & Landmark mistake. \\
    \cline{3-5}
    &  & thumb-pinky & 0.244 & MediaPipe's mistake for occluded fingers. \\
    \cline{2-5}
    & thumb down & thumb-index & 0.223 & Landmark mistake. \\
    \cline{2-5}
    & thumb backward & thumb-index & 0.249 & Same as above. \\
    \cline{2-5}
    & thumb forward & thumb-index & 0.186 & Same as above. \\
    \hline
  \end{tabular}
  }
\end{table*}

\clearpage

\subsection{Pseudocode for description rules}
\label{apx:pseudocode-rules}

\newcommand{\customsize}{\fontsize{9pt}{9pt}\selectfont}

\vspace{-3mm}

\begin{algorithm}
\caption{Flexion of a finger}
\customsize{
\begin{algorithmic}[1]
\Procedure{Flexion}{$finger, thresholdLow, thresholdHigh$}
\State $curl \gets 0$
\If {$finger$ is thumb}
\State $curl \gets Angle(\overrightarrow{\text{MCP-IP}}, \overrightarrow{\text{IP-TIP}})$
\Else
\State $curl \gets Angle(\overrightarrow{\text{MCP-PIP}}, \overrightarrow{\text{PIP-DIP}})$
\State $curl \gets curl + Angle(\overrightarrow{\text{PIP-DIP}}, \overrightarrow{\text{DIP-TIP}})$
\EndIf
\If {$curl \leq thresholdLow$}
\State \textbf{return} $straight$
\ElsIf {$curl \geq thresholdHigh$}
\State \textbf{return} $bent$
\Else
\State \textbf{return} $unsure$
\EndIf
\EndProcedure
\end{algorithmic}
}
\end{algorithm}

\vspace{-5mm}

\begin{algorithm}
\caption{Proximity of two fingers}
\customsize{
\begin{algorithmic}[1]
\Procedure{Proximity}{$finger_1, finger_2, threshold_{Low}, threshold_{High}$}
\State $jointDis \gets 0$
\State $polylineF1 \gets Polyline(finger_1\ PIP, finger_1\ DIP, finger_1\ TIP)$
\State $polylineF2 \gets Polyline(finger_2\ PIP, finger_2\ DIP, finger_2\ TIP)$
\State $distance_1 \gets Distance(finger_1\ PIP, polylineF2)$
\State $distance_2 \gets Distance(finger_2\ PIP, polylineF1)$
\State $jointDis \gets jointDis + min(distance_1, distance_2)$
\State $distance_3 \gets Distance(finger_1\ DIP, polylineF2)$
\State $distance_4 \gets Distance(finger_2\ TIP, polylineF1)$
\State $jointDis \gets jointDis + min(distance_3, distance_4))$
\State $jointDis \gets jointDis / 3$
\If {$jointDis$ is less than $thresholdLow$}
\State \textbf{return} $adjacent$
\ElsIf {$jointDis$ is greater than $thresholdHigh$}
\State \textbf{return} $separated$
\Else
\State \textbf{return} $unsure$
\EndIf
\EndProcedure
\end{algorithmic}
}
\end{algorithm}

\vspace{-5mm} 

\begin{algorithm}
\caption{Contact of two fingers}
\customsize{
\begin{algorithmic}[1]
\Procedure{Contact}{$finger_1, finger_2, threshold_{Low}, threshold_{High}$}
\State $distance \gets Distance({finger_1\ TIP, finger_2\ TIP})$
\If {$distance \leq thresholdLow$}
\State \textbf{return} $contact$
\ElsIf {$distance \geq thresholdHigh$}
\State \textbf{return} $not contact$
\Else
\State \textbf{return} $unsure$
\EndIf
\EndProcedure
\end{algorithmic}
}
\end{algorithm}

\vspace{-5mm}

\begin{algorithm}
\caption{Thumb Pointing Direction}
\customsize{
\begin{algorithmic}[1]
\Procedure{Contact}{$thumb, threshold$}
\If {$thumb$ is not straight}
\State \textbf{return} $unsure$
\Else
\State $Thumb \gets \overrightarrow{\text{MCP-TIP}}$
\State $minAngle \gets +\infty$
\State $ThumbDir \gets None$
\For {$dir$ in $[down,up]$}
\State $angle \gets Angle(Thumb, dir)$
\If {$angle$ is less than $minAngle$}
\State $minAngle \gets angle$
\State $ThumbDir \gets dir$
\EndIf
\EndFor
\If {$minAngle$ is greater than $threshold$}
\State \textbf{return} $unsure$
\Else
\State \textbf{return} $ThumbDir$
\EndIf
\EndIf
\EndProcedure
\end{algorithmic}
}
\end{algorithm}

\begin{algorithm}
\caption{Palm orientation}
\customsize{
\begin{algorithmic}[1]
\Procedure{PalmOrientation}{$hand, threshold$}
\State $PalmVec1 \gets \overrightarrow{pinkyMCP, indexMCP}$
\State $PalmVec2 \gets \overrightarrow{wrist, middleMCP}$
\If {$hand$ is left hand}
\State $PalmOriVec \gets PalmVec1 \times PalmVec2$
\ElsIf {$hand$ is right hand}
\State $PalmOriVec \gets PalmVec2 \times PalmVec1$
\EndIf
\State $minAngle \gets +\infty$
\State $PalmOri \gets None$
\For {$dir$ in $[right,left,down,up,outward,inward]$}
\State $angle \gets Angle(PalmOriVec, dir)$
\If {$angle$ is less than $minAngle$}
\State $minAngle \gets angle$
\State $PalmOri \gets dir$
\EndIf
\EndFor
\If {$minAngle$ is greater than $threshold$}
\State \textbf{return} $unsure$
\Else
\State \textbf{return} $PalmOri$
\EndIf
\EndProcedure
\end{algorithmic}
}
\end{algorithm}

\clearpage

\subsection{Gesture State Matrix Details}
\label{apx:rule_matrix}

The matrix generated by the rule-based module is subsequently interpreted by Gesture Description Agent's gesture summary description generation module, and the result is provided to the Gesture Inference Agent for further analysis. This matrix encapsulates hand pose and movement status across two distinct channels, each offering a different dimension of gesture representation. This pose-movement split is proven to promote Gesture Description Agent's performance, avoiding omitting important characteristics or overemphasizing certain aspect.

\paragraph{Channel 1: Hand Pose}
The first channel is a 2D array comprising 19 rows and \(T\) columns, where \(T\) denotes the number of time steps, with each step representing 0.2 seconds. The rows are indexed starting from 1 and detail the following aspects:

\begin{itemize}
    \item Rows 1-5 correspond to finger flexion for the thumb, index, middle, ring, and pinky fingers, respectively. The values are encoded as 1 (straight), 0 (between straight and bent), and -1 (bent), describing the extent of finger flexion.
    \item Rows 6-8 represent finger proximity for adjacent finger pairs (index-middle, middle-ring, ring-pinky) with similar encoding scheme. The aim is to indicate how closely each finger is to its neighbor.
    \item Rows 9-12 detail thumb fingertip contact with the fingertips of the other fingers, again using similar value encoding.
    \item Row 13 specifies the pointing direction of the thumb, with 1 (upward), -1 (downward), and 0 (no specific direction or unknown when thumb is bent).
    \item Rows 14-19 are dedicated to palm orientation, indicating the direction the palm faces from the user's perspective. A specific orientation is marked by a single row set to 1 among these rows, representing left, right, down, up, inward, and outward directions. All rows equal to 0 means no specific direction can be identified.
\end{itemize}

\paragraph{Channel 2: Hand Movement}
The second channel consists of a 2D array with 2 or 3 rows (depending on whether we can extract hand position in 3d space or 2d space) and \(T\) columns. On each time step, the vector corresponds to the geometric center of the hand at this time:

\begin{itemize}
    \item Row 1 tracks the horizontal position (0 for leftmost to 1 for rightmost), where increasing values suggest movement from left to right.
    \item Row 2 follows the vertical position (0 for bottommost to 1 for topmost), where increasing values suggest movement from down to up.
\end{itemize}

To gauge movement magnitude, the hand's width is also provided. For example, a hand width of 0.05 with a rightward movement of 0.05 in the array suggests a displacement of approximately one hand width.

This detailed matrix representation ensures a comprehensive and nuanced understanding of hand gestures, facilitating advanced processing and interpretation in gesture recognition systems.

\clearpage 

\section{Detailed Experiment Setting}
\label{appendix:exp_setting}

This section outlines the experimental setup for two experiments carried out within this research.

\subsection{Experiment 1: Augmented Reality-Based Smart Home IoT Control}
\label{appendix:exp_setting_home}

\begin{figure*}[htbp]
  \centering
  \includegraphics[width=\textwidth]{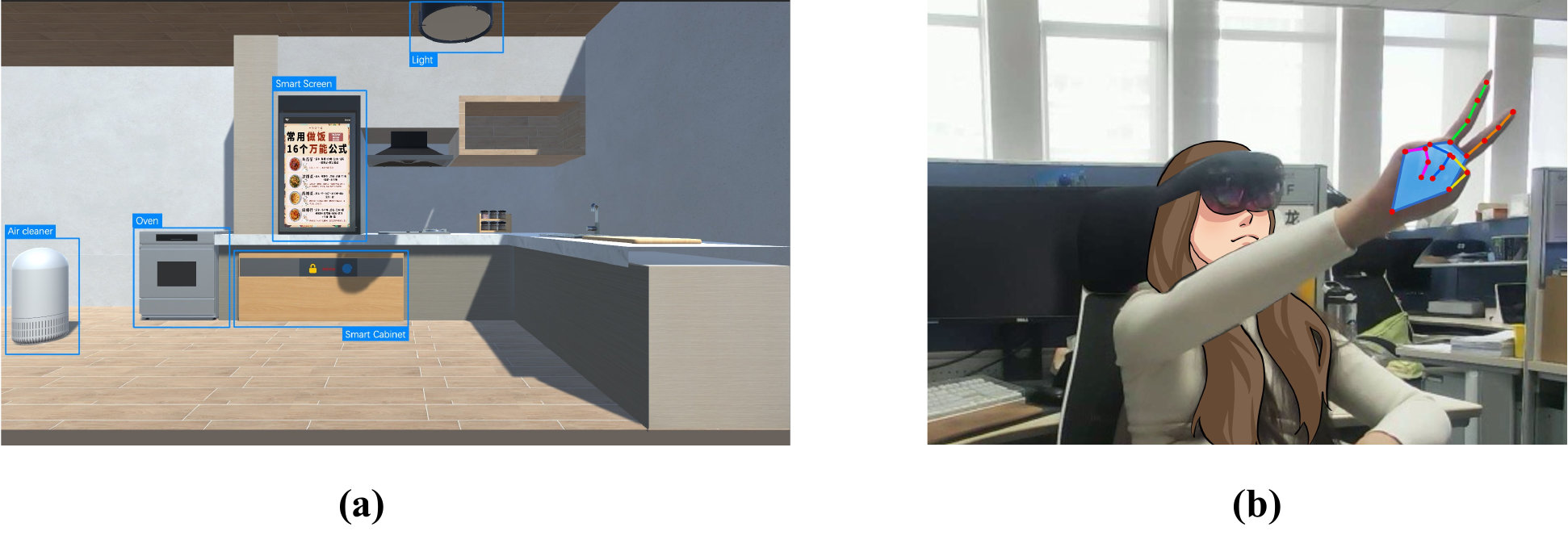}
  \caption{The experimental platform utilized in the smart home scenario. (a) IoT device control interface, simulated using Unity. (b) User wearing the Hololens and performing gestures with the right hand for device control.}
  \label{fig:appendix_exp_interface_home}
\end{figure*}

The first experiment focuses on controlling IoT devices within a smart home environment through augmented reality. The setup simulates a scenario where users interact with various home devices using gesture controls.

As illustrated in Figure~\ref{fig:appendix_exp_interface_home}, the experimental platform integrates an augmented reality interface for IoT device control within a home setting.

The experimental setup encompasses a variety of device functions and user tasks to mimic real-world interactions with a smart home environment. Details for this experiment are presented in Table~\ref{tab:home_device_function} for device functions and Table~\ref{tab:smart_home_info} for tasks and external contexts. 
The smart home scenario encompasses a total of 18 functions across 5 devices, offering a comprehensive assessment of gesture-based control in an augmented reality context.

\begin{table}[t!]
  \caption{Task Instructions and External Context in the Smart Home Scenario}
  \label{tab:smart_home_info}
  \centering
  \resizebox{\columnwidth}{!}{
  \begin{tabular}{|p{4cm}|p{9cm}|}
    \hline
    \textbf{Task Instruction} & \textbf{External Context}\\
    \hline
    Unlock the Smart Carbinet. & \makecell[l]{[``It is 7:00 PM now.'',\\ \ ``The child lock on this cabinet supports fingerprint unlocking.'']} \\
    \hline
    Increase the brightness of the light. & [``It is 7:05 PM now.''] \\
    \hline
    Show the next recipes on the smart screen. & [``It is 7:12 PM now.''] \\
    \hline
    Open the oven. & \makecell[l]{[``It is 7:14 PM now.'', \\ \ ``Recipe instructions: now you need to open the oven'']} \\
    \hline
    Open the air cleaner. & \makecell[l]{[``It is 7:20 PM now.'', \\ \ ``The air purifier's sensor detected that the current environment \\ has heavy cooking fumes.'']}\\
    \hline
    Set a timer on the smart screen. & \makecell[l]{[``It is 7:30 PM now.'', \\ \ ``Recipe instructions: \\ now you need to cook on high heat for five minutes.'']} \\
    \hline
    Switch input source of the smart screen to the smart bell. & \makecell[l]{[``It is 7:32 PM now.'', \\ \ ``The doorbell is ringing.'']} \\
    \hline
    Make a phone call throught the smart screen. & \makecell[l]{[``It is 7:33 PM now.'', \\  ``Just now, it was the deliveryman delivering goods;\\ the owner of the goods is the user's roommate, Mark.'']} \\
    \hline
  \end{tabular}
  }
\end{table}

\begin{table*}[t!]
  \centering
  \caption{Device Functions in Smart Home Scenario (Totally 18 Functions in 5 Devices)}
  \label{tab:home_device_function}
  \resizebox{\columnwidth}{!}{
  \begin{tabular}{|l|l|}
  \hline
  \textbf{Device Name} & \textbf{Function Name} \\
  \hline
  \multirow{3}{*}{Light} & On / Off \\
  \cline{2-2}
  & Brightness Control \\
  \cline{2-2}
  & Mode Switch (Task Lighting / Morning Lighting / Accent Lighting) \\
  \hline
  \multirow{3}{*}{Smart Cabinet} & Child Lock Activated / Deactivated \\
  \cline{2-2}
  & Temperature Control \\
  \cline{2-2}
  & Humidity Control \\
  \hline
  \multirow{5}{*}{Smart Screen} & On / Off \\
  \cline{2-2}
  & Switch Recipes (Recipe 1 / Recipe 2 / Recipe 3) \\
  \cline{2-2}
  & Switch Input Source (Smart Screen / Smart Doorbell / IPad Video) \\
  \cline{2-2}
  & Phone Call \\
  \cline{2-2}
  & Settable Timer \\
  \hline
  \multirow{4}{*}{Oven} & On / Off \\
  \cline{2-2}
  & Temperature Control \\
  \cline{2-2}
  & Self Cleaning On / Off \\
  \cline{2-2}
  & Mode Switch (Bake Mode / Convection Roast / Bottom Heat Only / Keep Warm / Energy Efficiency) \\
  \hline
  \multirow{3}{*}{Air Cleaner} & On/Off \\
  \cline{2-2}
  & Airflow Speed Control \\
  \cline{2-2}
  & Mode Switch (Strong / Silent / Custom) \\
  \hline
  \end{tabular}
  }
\end{table*}

\subsection{Experiment 2: Online Video Streaming on PC}
\label{appendix:exp_setting_video}

\begin{figure*}[t!]
  \centering
  \includegraphics[width=\textwidth]{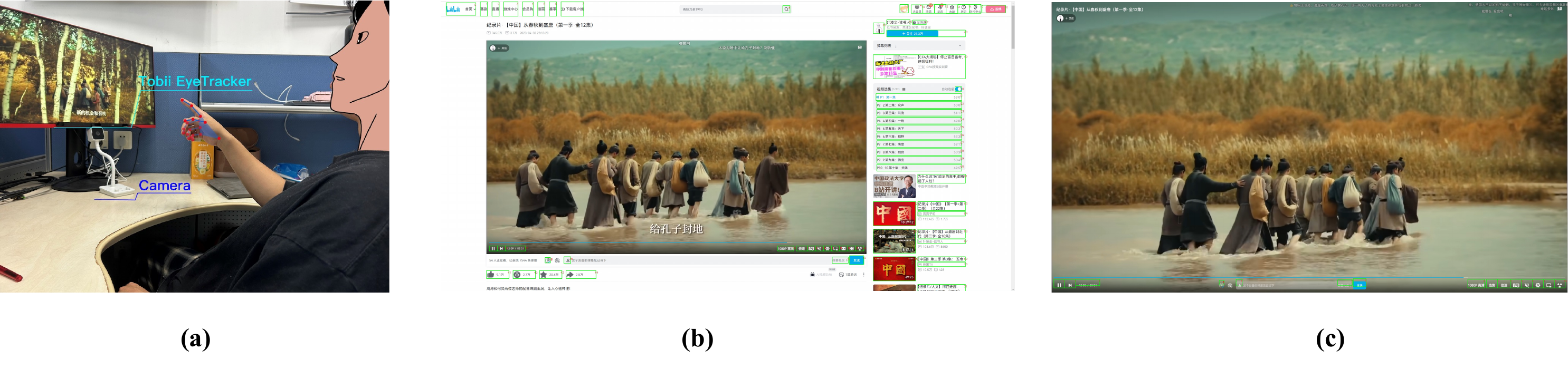}
  \caption{The experimental platform utilized in the video streaming scenario. (a) A user was watching the video and performing gesture to control it. (b) and (c) are the video interface and the function distribution. }
  \label{fig:appendix_exp_interface_video}
\end{figure*}

The second experiment focuses on user interaction with online video content on a PC monitor. 
To ensure familiarity with the website's interface among participants, we selected a highly popular video website for the experiment. 
Figure~\ref{fig:appendix_exp_interface_video} illustrates the setup of our experimental platform.

Detailed descriptions of the tasks, including the number of functions and external context information for the video streaming environment, are provided in Table~\ref{tab:video_streaming_info}. 
Additionally, the list of functions available for tasks 4, 5, and 6 is presented in Table~\ref{tab:video_function_list_2}, while the function list for the remaining tasks is shown in Table~\ref{tab:video_function_list_1}. 
The variation in the function list is attributed to task 3, which involves full-screening the video page, resulting in a reduced number of available functions.

\begin{table}[t!]
  \caption{Task Instructions, Function Numbers and External Context in Video Streaming Scenario}
  \label{tab:video_streaming_info}
  \resizebox{\columnwidth}{!}{
  \begin{tabular}{|p{4cm}|p{3cm}|p{6cm}|}
    \hline
    \textbf{Task Instruction} & \textbf{Function Numbers} & \textbf{External Context}\\
    \hline
    Turn up the volume. & 66 & [``It is 8:01 PM now.''] \\
    \hline
    Drag the progress bar forward. & 66 &  \makecell[l]{[``It is 8:02 PM now.'', \\ \ ``The user has watched the earlier part of \\ this video.'']}\\
    \hline
    Enter full screen mode. & 66 &  [``It is 8:04 PM now.'']\\
    \hline
    Pause the video. & 17 &  \makecell[l]{[``It is 8:15 PM now.'', \\ \ ``Right now, the user's phone is actively \\receiving an incoming call.'']} \\
    \hline
    Resume the video. & 17 &  \makecell[l]{[``It is 8:17 PM now.'', \\ \ ``The user hung up the phone'']} \\
    \hline
    Exit full screen mode. & 17 &   [``It is 8:48 PM now.'']\\
    \hline
    Like the video. & 66 &   [``It is 8:49 PM now.'']\\
    \hline
    Go to the next episode. & 66 &  [``It is 8:50 PM now.'']\\
    \hline
  \end{tabular}
  }
\end{table}

\begin{table}[t!]
\centering
\caption{Video Scenario Function List in Task 4, 5, 6}
\label{tab:video_function_list_2}
\begin{tabular}{|c|l|c|l|}
\hline
\textbf{ID} & \textbf{Name}                 & \textbf{ID} & \textbf{Name}                \\ \hline
0           & VideoProgressBarUpdate        & 9           & SelectEpisode                \\
1           & PlayPauseButton               & 10          & ChangePlaybackSpeed          \\
2           & NextButton                    & 11          & SubtitleControl              \\
3           & SeekTimeUpdate                & 12          & VolumeControl                \\
4           & ToggleDanmakuDisplay          & 13          & VideoSettingsMenu            \\
5           & DanmakuToggle                 & 14          & PictureInPictureToggle       \\
6           & DanmuEtiquetteHint            & 15          & ToggleFullscreen             \\
7           & SendMessageButton             & 16          & VideoPlayArea\\
8           & VideoQualitySelection         &             &                              \\ \hline
\end{tabular}
\end{table}

\section{System Cost for LLM Use}
Our system employs the OpenAI API\footnote{https://openai.com/pricing\#language-models} \texttt{gpt-4-1106-preview} for experiments, which is one of the best performance LLM and achieving near-human-level common sense and reasoning. 
The cost of each run is determined by the total token count.

For each gesture, GestureGPT consumes an average of \(38785\) tokens (\(\text{SD} = 10432\)) for input and an average of \(3443\) tokens (\(\text{SD} = 1339\)) for output, spanning \(6.08\) rounds of conversation (\(\text{SD} = 0.85\)). This results in a cost of \(\$0.389\) per gesture. 

Although costs are currently elevated due to the premium on model resources and extensive token requirements, 
we anticipate a reduction in expenses as LLM technology continues to evolve. 

\section{Data Samples and Prompts Utilized in GestureGPT}
\label{appendix:prompt}

GestureGPT utilizes a triple-agent collaborative architecture, with each agent guided by its own specific prompt to steer the behavior of the LLM. The gesture data in GestureGPT also undergoes several format transformations. To facilitate better utilization by the community, we will make our prompts and some corresponding data samples publicly available. 
Due to the length of the system prompt, it has been archived in a GitHub repository. The full prompt can be accessed at \url{https://github.com/studyzx/GestureGPT_ISS}.

\bibliographystyle{ACM-Reference-Format}
\bibliography{reference_new}

\end{document}